\documentclass[journal,twoside,web]{ieeecolor2}
\usepackage{generic}
\usepackage{cite}
\usepackage{amsmath,amssymb,amsfonts}
\usepackage{algorithmic}
\usepackage{graphicx}
\usepackage{textcomp}
\usepackage{comment}
\usepackage{mathtools}
\usepackage{siunitx}
\usepackage{makecell}
\usepackage{booktabs}
\usepackage[ruled,linesnumbered,noend]{algorithm2e}
\usepackage[switch]{lineno}
\def\BibTeX{{\rm B\kern-.05em{\sc i\kern-.025em b}\kern-.08em
    T\kern-.1667em\lower.7ex\hbox{E}\kern-.125emX}}
\newcommand{\TIIPS}{TI$^{2}$PS}
\newcommand{\added}[1]{#1}
\newcommand{\removedtext}[1]{}
\newcommand{\replaced}[2]{#2}

\begin{document}
%\linenumbers
\title{TI$^2$PS: A Topology-Informed Inverse Design Framework for Stochastic Multicellular Pattern Formation}
\author{
  Kenji Komiya$^{*1}$, Andrew Kailiang Jin$^{*1,*2}$, Ryo Nishikimi, and Kunio Kashino
  \thanks{Manuscript received xx xx xxxx; revised xx xx xxxx.}%
  \thanks{This article is an extended version of \cite{jin2025topology}.}
  \thanks{Kenji Komiya, Ryo Nishikimi, and Kunio Kashino are with NTT, Inc., Kanagawa 2430124 Japan (e-mail: kenji.komiya@ntt.com).}%
  \thanks{Andrew Jin is with the Georgia Institute of Technology, Atlanta, GA 30332 USA (e-mail: ajin40@gatech.edu).}%
  \thanks{*1 These authors contributed equally to this work.}%
  \thanks{*2 This work was conducted during an internship project at NTT, Inc.}%
}

\maketitle

\begin{abstract}
\textcolor{cyan}{\textit{Objective:}}
This study proposes a novel framework to estimate parameters for reproducing target multicellular patterns using an agent-based model (ABM). 
Two major challenges in multicellular ABMs are estimating cell-level parameters (agent-specific variables) and quantitatively evaluating the topological characteristics of multicellular arrangements under stochastic cell proliferation and death.
To address these challenges, we integrate two approaches: Betti vectors and inverse surrogate modeling.
The Betti vectors obtained through topological data analysis
 can consistently represent features of a wide range of multicellular spatial configurations.
The inverse surrogate modeling 
 enables direct inference of the corresponding ABM parameters from the target patterns.
We validated the proposed framework using zebrafish pigment pattern formation, 
 a representative model of pattern formation driven by multicellular interactions.
The results demonstrate that our framework successfully estimates ABM parameters and outperforms conventional methods such as PointNet++. 
\replaced{
Notably, the proposed method, which used only 10\% of the training data, achieved higher accuracy than conventional methods that used 100\% of the data.
}
{
Notably, the proposed method, which used only 10\% of the training data, outperformed PointNet++, which used 100\% of the data, across all evaluation metrics.
}
\removedtext{
Additionally, when we applied the framework to mutant zebrafish pigment patterns, 
 we estimated parameters with limited similarities to target patterns. 
This discrepancy suggests that the framework may also serve as a detection tool for identifying missing or unknown mechanisms in the underlying ABM or biological system.
}
\end{abstract}


\section{Introduction}
{\bf Significance of elucidating mechanisms of multicellular behavior} -
%%% Suggested Multicellular Behavior Introduction:
Multicellular pattern formation is a biological process in which cells self-organize into spatially distinct structures. This process is essential for the development of complex tissues and organs, as well as during tissue repair and regeneration. These complex patterns arise from the coordination of cell-cell interactions and environmental cues. Many diseases, including some cancers and genetic disorders, yield irregular patterns from abnormal cellular behavior that result in the dysfunction of these multicellular interactions. A goal of biomedical researchers is to understand the mechanisms that control multicellular pattern formation, and to leverage these principles to understand disease pathology and treatments, and to create engineered 
tissues~\cite{bianco2001stem,hollister2005porous,levenberg2005engineering}, artificial organs~\cite{hofer2021engineering,mandrycky20163d,murphy20143d}, and other synthetic biological systems~\cite{cianchetti2018biomedical,lutolf2009designing,bruggeman2007nature}.
%%%

{\bf Vitalization of computer simulation of cellular behavior} - 
% 細胞のふるまいのシミュレーションは大事．
%The modeling of cellular behavior is essential for predicting and controlling biological development and tissue organization. 
% 細胞のふるまいのシミュレーションには色々なアプローチがある．
Various computational modeling approaches have been developed to understand the mechanisms that underlie multicellular pattern formation,
 including reaction-diffusion systems~\cite{HusarOGYSFBKS24,HepsonYA21,ArjunanMIT20,ChenLWC17},
 mechanical models~\cite{BoudonCAGHBTG15,GolabSSM14,ZhaoNKCTL13},
 and agent-based models (ABMs)~\cite{CortiCMBCMC22,ChenV19}.
% ABS一般の話
Agent-Based Models (ABMs) are multi-scale, computational models that represent each cell as an information processing unit or “agent”. Each agent is governed by a set of rules that dictate its behavior, which may be dependent on interactions with the simulation environment or other agents. This approach is particularly well-suited for understanding multicellular behavior as a "bottom-up" model because it enables the study of emergent pattern formation through explicit representation of cellular interactions.
% ABSを細胞移動のシミュレーションに適応することのメリットの話

\begin{figure}[t]
  \includegraphics[ width=\columnwidth]{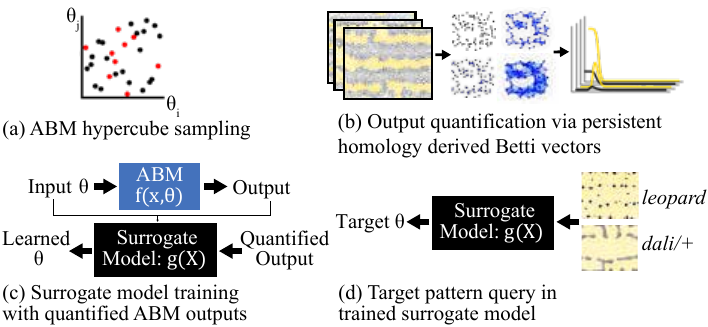}
  \caption{Proposed agent-based model (ABM) optimization workflow. (a) A training and testing dataset of the desired ABM is generated through hypercube sampling. (b) Cellular ABM outputs can be generically quantified using topological data analysis, resulting in a series of Betti vectors that represent the topological features of the ABM output. (c) Surrogate model training and testing using quantified model outputs (Betti vectors) and known input parameter values. (d) Target patterns can be queried into the trained model to elucidate parameter combinations that may yield similar patterns in simulation.}
  \label{fig_abstract_fig}
\end{figure}

{\bf Current challenges for computational modeling of cellular behavior} - 
%In practice, there have been two difficulties in using ABMs to model cellular behavior in nature: 
Although ABMs are promising for studying multicellular behavior, there are two main challenges for effective and efficient implementation. (See also the excellent survey papers, e.g., \cite{glen2019agent} and Section $4$ in \cite{wang2015simulating}.)
\begin{itemize}
\item {\bf Difficulty in estimating cellular-level (agent-specific) variables} - One of the typical drawbacks of ABM has long been the difficulty of estimating agent-specific parameters~\cite{monti2023learning}. A common approach to this challenge is to explore a parameter space in a grid search manner~\cite{an2012modeling}. Unfortunately, it is often impractical to explore high-dimensional parameter spaces within a limited time because each ABM simulation of individual cells and their local interactions is computationally intensive.
\item {\bf Difficulty in handling stochastic proliferation and death of cells} - In order to handle cell proliferation and death within the ABM framework, it is necessary to have varying model complexity (i.e., number of agents)~\cite{wang2015simulating}, which significantly affects the difficulty and instability of estimating agent-specific parameters.
\end{itemize}
Therefore, there has been a continuing demand for methods to resolve these two interdependent issues. % in the modeling of cellular behavior. % using ABMs.

\begin{comment}
One of the challenges in using ABMs is finding appropriate parameters 
 that allows the models to regenerate observed biological patterns~\cite{monti2023learning} (i.e., the number and arrangement of cells)
% To regenerate biological phenomena with ABMs, we need to find appropriate parameters for the model that can regenerate biological patterns.
% However, ABM faces difficulties in finding these parameters.
% The parameters used in ABM are often difficult or impossible to determine experimentally, 
 because cellular-scale information on the parameters is unavailable or inaccessible in vivo, even in vitro.
A common approach to this challenge is to explore a parameter space in a grid search manner~\cite{an2012modeling}.
However, it is often impractical to explore high-dimensional parameter spaces within limited time
 because each run of ABMs simulating individual cells and their local interactions is computationally intensive.
\end{comment}

{\bf Our key strategy} - As a method capable of simultaneously addressing these two interdependent issues, we propose a new framework for ABM that combines the principles of \emph{surrogate modeling} and \emph{topological data analysis} (TDA)~\cite{chazal2021introduction,rabadan2019topological,SkafL22} (Fig. \ref{fig_abstract_fig}).
For the first key feature, we introduce an inverse surrogate module that estimates the parameters of the biological system simulator from the simulator's outcomes, forming biological patterns.
For the second key feature, we introduce the TDA-inspired module, which provides a means to assess the global structure of patterns. This allows for meaningful comparison even when the number, position, or scale of features differs between the simulation and the target.
We refer to this approach as the Topology-Informed Inverse Parameter Surrogate (TI$^{\mathbf{2}}$PS) framework~\cite{jin2025topology}, emphasizing the integration of both Topology and Inverse parameterization in a surrogate modeling pipeline.

To evaluate the feasibility of the proposed framework, 
 we applied the TI$^{\mathbf{2}}$PS approach to simulations of zebrafish pigment pattern formation in development~\cite{volkening_modelling_2015}.
This zebrafish pigment system provides an ideal testbed, utilizing two interacting cell types to produce characteristic spatial patterns governed by fundamental cellular behaviors — migration, division, and death~\cite{volkening_modelling_2015}.
%As surrogate models, 
We implemented %two different machine learning architectures:
 the inverse surrogate model based on 
 a generalized linear model (GLM) and
 a multilayer perceptron (MLP). 
 % with two different activation functions.
Furthermore, we applied the trained inverse surrogate model to estimate parameters from mutant zebrafish pigment patterns. 
 
 \begin{comment}
Both models were trained to minimize the difference between the true simulation parameters and the parameters inferred by the GLM.
Inferred parameters successfully regenerate target patterns.\
 Furthermore, we applied the trained inverse surrogate model to estimate parameters from mutant zebrafish pigment patterns. 
The \textit{dali/+} mutant is an artificially induced variant characterized by distorted stripes, consisting of regularly interrupted stripes with irregular orientations~\cite{delcourt2018individual}. 
Its causative gene remains unidentified; however, previous studies have suggested that the mutation affects cell motility and cell-cell interactions~\cite{inoue2014tetraspanin}. 
The \textit{leopard} mutant is another artificially induced variant, characterized by a spotted black pigment pattern~\cite{delcourt2018individual}. 
This mutation is caused by a defect in the \textit{cx41.8} gene~\cite{watanabe2006spot}, which encodes Connexin 41.8, a gap junction protein essential for cell-cell communication during pigment pattern formation. 
The simulations using the parameters inferred by the surrogate model reproduced the observed mutant patterns with limited concordance, suggesting that our surrogate model method can also highlight when a proposed ABM lacks mechanisms necessary to fully capture these phenotypes. 
These results confirm that TI$^{\mathbf{2}}$PS not only enables efficient inverse inference but also serves as a tool for detecting missing biological mechanisms in ABMs.
\end{comment}
 %Related worksを入れる（最後）
\section{Related Work}
This paper focuses on the intersection of ABMs, surrogate models, and TDA, which are recent popular tools used to shed light on the principles of cellular behavior. In this section, we clarify the novelty of this paper in terms of recent developments regarding these three tools.

{\bf Multiscale extension of ABMs} - Recent reviews have emphasized that ABMs are promising tools for linking macroscopic tissue-level phenomena with microscopic cellular dynamics~\cite{norton2019multiscale,wang2015simulating,zhang2009multiscale,an2016exploring}.
A key challenge in this integration is inferring cell-level behavioral rules from observable tissue-level patterns.

{\bf Affinity of ABMs and surrogate models} - 
The combination of ABM and surrogate models is an especially active area that has been attracting a lot of attention in recent years~\cite{angione2022using,fabiani2024task,zhang2020validation}.
In general, surrogate modeling aims to use simplified models to approximate ABM's agent-specific variables efficiently.
However, a straightforward application to its cellular behavior is not appropriate.
Due to the stochastic nature of cell proliferation, death, and other biological processes, the number and arrangement of cells may vary across simulations, making one-to-one correspondence between simulated and reference patterns difficult.

{\bf Affinity of ABMs and TDA} - 
The combination of topological data analysis with ABM has become popular in recent years
~\cite{mcguirl2020topological,nguyen2024quantifying}. In general, TDA is a set of mathematical techniques that extract structural features (connectivity, loops, and voids) from spatial data~\cite{chazal2021introduction,rabadan2019topological,SkafL22}.
Instead of relying on point-wise correspondence, TDA provides a means to assess the global structure of patterns, allowing for meaningful comparison even when the number, position, or scale of features differs between the simulation and the target.
However, the incorporation of TDA itself does not have the ability to directly alleviate the difficulty of estimating ABM's agent-specific variables.

{\bf Novelty of our approach} - As a nexus of the above three trends in ABM, this paper proposes a new way to make the estimation of cell-level (agent-specific) variables tractable while capturing the global structure of cellular behavior by introducing a surrogate model and a TDA mechanism to ABM in an explicit way.

{\bf \replaced{Conevntional approaches}{Conventional approaches}} - 
Multicellular patterns consist of the positions of each cell type within a tissue. 
In biological systems, the number of cells in a pattern is not constant even under the same conditions because of natural randomness. 
Furthermore, the specific order of cells in a dataset does not provide meaningful information. Therefore, methods for this problem must work regardless of the number of cells or their order.
Pooling layers solve these issues by summarizing information from a variable number of points into a single vector of a fixed size. 
For example, max-pooling in a high-dimensional space extracts the maximum value for each feature across all points. 
This ensures that the model can handle different cell counts and remains independent of the cell ordering.
One basic approach is to directly pool coordinates. 
Cell pattern data are often collected from a specific rectangular region within a tissue. 
Using maximum, minimum, and average pooling on each coordinate extracts information about the nearest cell position to the corner and the distribution biases of each cell type within that region.
There are also advanced pooling approaches used for general point cloud tasks. 
PointNet~\cite{qi2017pointnet} is a well-known method that uses global max pooling in a high-dimensional space to connect spatial distributions to model outputs. This method enables the efficient extraction of global features such as edges and corners from a set of points. 
PointNet++~\cite{qi2017pointnet++} is an improved version that captures both local and global geometric features through hierarchical grouping and pooling. 
This study uses these two methods as strong baselines because they are standard and effective approaches for processing point-based data.
\added{
Several previous studies have also applied topological data analysis to biological pattern formation.
For example, McGuirl et al.~\cite{mcguirl2020topological} analyzed wild-type and mutant zebrafish pigment patterns using topological descriptors.
Such descriptors are independent of the order of points and can characterize spatial pattern structures.
However, their objective was pattern characterization rather than inverse surrogate modeling, and their framework was not designed to estimate ABM parameters from stochastic multicellular point-cloud patterns.
Therefore, although this work is closely related to our motivation, it cannot be directly used as a baseline for the parameter estimation task considered in this study.
}

\section{Method} \label{sec:method}

%Preliminary: Agent Based Model for Zebrafish Stripe
%Proposed: Inverse surrogate modeling 
%Pattern metric 

This section specifies the problem
 of estimating the parameters of a biological system simulator
 from observed biological patterns,
 and describes the proposed inverse surrogate modeling approach to this problem.

\subsection{Problem Specification}

Our goal is to construct the inverse surrogate model
 that takes as input an observed biological pattern
 and outputs the parameters of a biological system simulator.
The input biological pattern
 is a set of cell positions $\mathbf{X} = \left\{\mathbf{X}^c\right\}_{c\in\mathcal{C}}$
 on a two-dimensional plane,
 where $\mathcal{C}$ is a set of cell types.
Each $\mathbf{X}^c = \left\{\mathbf{x}^c_i\right\}_{i=1}^{N_c}$
 is a set of cell positions of type $c$,
 where $N_c$ is the number of cells of type $c$ and
 $\mathbf{x}^c_{i}\in\mathbb{R}^2$ represents a two-dimensional coordinate
 of an $i$-th cell of type $c$.
The output of our model
 is simulator's parameters $\boldsymbol{\Theta}=\{\boldsymbol{\theta}^{cc'}\}_{c,c'\in\mathcal{C}}$,
 where each parameter set $\boldsymbol{\theta}^{cc'}=\{\theta^{cc'}_d\}_{d=1}^{D_\theta}\in\mathbb{R}^{D_\theta}_{>0}$ is indexed by the pair of cell types $c$ and $c'$
 to represent the relationships between the same or different cell types.

\begin{comment}
\begin{figure}[t]
\centerline{\includegraphics[width=0.65\columnwidth]{fig/rips_filtration.pdf}}
\caption{Schematic illustration of Vietoris-Rips filtration and simplex construction.
The black points are the centers of the blue balls, and
 each ball is centered at a cellular position.
(a) Each ball center %without intersection with the other balls
    represents a 0-simplex (point). % radius $\epsilon_i / 2$ (0-simplex). 
(b) Two intersecting balls represent a 1-simplex (edge) between their center points. 
(c) Three mutually intersecting balls represent a 2-simplex (filled triangle) with vertices at their center points.
}
\label{fig_rips}
\vspace{-8.0mm}
\end{figure}
\end{comment}

\subsection{Betti Vector}\label{sec:betti}
The Betti vector can represent the topological features of the cellular point cloud
 and is derived from the theory of persistent homology, 
 which can capture spatial patterns across multiple scales.
For each cell type $c \in \mathcal{C}$,
 we compute the Betti vector according to the following steps.

\textbf{Step 1: Construction of Vietoris-Rips Filtration} - 
We consider a set of balls with radius $\epsilon / 2$,
 centered at the positions of $M$ cells.
Here, $\epsilon \in \mathbb{R}_{\geq 0}$ is called a filtration value, 
 $M$ is the number of cells, and 
 each ball center is indexed sequentially from $1$ to $M$.
Under this condition, we construct a simplicial complex $X(\epsilon)$, i.e., a set of simplices
 based on the intersections of these balls.
A simplex is defined by a set of center indices,
 where $k$-simplex consists of $k+1$ indices and forms a specific geometric structure.
For example, the 0-simplex, 1-simplex, and 2-simplex represent
 a point, a line segment connecting two points, and a filled triangle connecting three points, respectively. 
First,
 we add all indices of ball centers (0-simplices) to $X(\epsilon)$.
Next, if two balls intersect,
 a pair of their indices (1-simplex) is added to $X(\epsilon)$.
Finally, if three balls intersect each other,
 a tuple of their center indices (2-simplex) is added to $X(\epsilon)$.

\textbf{Step 2: Computation of Persistent Homology} - 
We observe the changes that occur in the topological features of $X(\epsilon)$
 with the gradual increasing of $\epsilon$.
In particular,
 we focus on the \textit{birth} and \textit{death}
 of two topological features: a connected component and a loop.
The connected component is a $0$-degree topological features.
At $\epsilon=0$,
 each $M$ ball center forms an independent connected component.
As the value of $\epsilon$ increases,
 adjacent components merge to form a larger connected component.
The loop is a $1$-degree topological feature
 and is a cycle formed by multiple center points connected with 1-simplices (i.e., line segments).
When $\epsilon = 0$,
 no loops exist.
However, as the value of $\epsilon$ increases, new loops emerge, 
 while existing ones may disappear when they are filled by 2-simplices (i.e., triangles).
Based on this observation,
 we can represent the gradual changes in the topological features of each degree $k\in\{0, 1\}$ as
\begin{align}
 \mathcal{P}^{k} = \Big\{ \left( b_j^{k}, d_j^{k} \right) \Big\}_{j=1}^{N_{k}},
\end{align}
where
 $N_{k}$ is the total number of $k$-degree topological features,
 $b_j^{k}$ is
 the value of $\epsilon$ when $j$-th $k$-degree topological feature appears, and 
 $d_j^{k}$ is
 the value of $\epsilon$ when it disappears.
 $\mathcal{P}^k$ is mathematically derived based on the theory of homology in topology
 \cite{bauer2021ripser}. 
 % we skip the explanation here.

\begin{comment}
我々は，Eを徐々に増やすと変化するX（E）に関するトポロジカル特徴を解析する．
特に，２つの特徴量，連結成分とループの”誕生”と”死亡”にフォーカスする．
まずは，連結成分について．
E=0のとき，M個の連結成分が既に存在（誕生）している．
そして，Eを徐々に増やしていくと，隣り合う２つの成分は併合し（死亡），１つの大きな連結成分が登場する（誕生）．
次は，ループについて．
E=0のとき，ループは存在しない．
そして，Eを徐々に増やしていくと，１単体で結ばれた複数の中心点がサイクルを形成する時，ループが誕生し，そのサイクルが２単体によって完全に埋められたときに，そのループは消滅する．

これらP0やP1はX(E)から導出できるが，その方法は数学的に複雑なので説明を割愛する．

Wwhen E = 0,
 there are M connected components, each of which corresponed to each cell point.
When E is gradually increased,
 adjacent two components merge (death) and
 larger one component appear (birth).

When E = 0,
 there is no loops.
A loop is born when a cycle is formed by connecting multiple points with 1-simplices,
 and dies when the cycle is completely filled by 2-simplices, thereby eliminating the corresponding hole.

The connected components
 are born when an isolated cell appears in the point cloud,
 and dies when two components merge as the filtration progresses with increasing $\epsilon_n$.

\end{comment}

\textbf{Step 3: Computation of Betti curves} - 
Using the set of birth-death pairs $\mathcal{P}^{k}$ obtained in Step 2, 
 we define the Betti curves,
 a function that returns the number of topological features (i.e., connected components and loops)
 alive at filtration value $\epsilon$ for each dimension $k$,
 as follows:
% For each dimension $k$ and filtration value $\epsilon$, 
%  the corresponding Betti curve $\beta^{k}(\epsilon)$ is calculated as
\begin{align}
 \beta^{k}(\epsilon) = \sum_{j=1}^{N_k}\mathbf{1}\left( b_j^{k} \leq \epsilon < d_j^{k} \right),
\end{align}
where
 $\mathbf{1}(\cdot)$ denotes the indicator function, 
 which equals $1$ if the condition given in parenthesis holds and $0$ otherwise. 
% Thus, $\beta^{0}(\epsilon = \epsilon_i)$ and $\beta^{1}(\epsilon = \epsilon_i)$
%  represent the number of connected components
%  and that of loops 
%  \textcolor{red}{that are alive at filtration value $\epsilon_i$}.

\begin{comment}
\begin{figure}[t]
\centerline{\includegraphics[width=0.67\columnwidth]{fig/persistent_homology.pdf}}
\caption{Schematic illustration of "birth" and "death" in persistent homology.
(a) Regarding the connected components, 
 \textit{birth} represents the appearance of isolated cells, and 
 \textit{death} occurs when two components merge as the filtration progresses. 
(b) Regarding the loops, 
 \textit{birth} represents the formation of a cycle, and 
 \textit{death} occurs when the cycle is completely filled.
}
\label{fig_persis}
\vspace{-6mm}
\end{figure}
\end{comment}

\newcommand{\concat}[0]{;\,}

\textbf{Step 4: Construction of Betti Vector} - 
% The Betti curves are concatenated across different homology dimensions and cell types to construct a fixed-dimensional feature vector for subsequent analysis.
Let $\beta^{k}_{c}(\epsilon)$ be the $k$-th Betti curve obtained from the set of cell positions of type $c$,
 the Betti vector is defined by 
\begin{align}
  \mathbf{v}^{k}_{c} = \begin{bmatrix}
      \beta^{k}_{c}(\epsilon_1) & \cdots & \beta^{k}_{c}(\epsilon_{N_E})
  \end{bmatrix},
\end{align}
where
 $\{\epsilon_i\}_{i=1}^{N_E}$ is a monotonically increasing sequence of filtration values,
 and $\beta^{k}_{c}(\epsilon_i)$ is the Betti number for the filtration value $\epsilon_i$.
Finally, by concatenating $\mathbf{v}^{k}_{c}$ across all degrees $k$ and all cell types $c$,
 we obtain the full Betti vector as
\begin{align}
 \mathbf{v} = \left[
    \mathbf{v}^0_{c_1} \concat\cdots \concat \mathbf{v}^0_{c_{|\mathcal{C}|}}\concat
    \mathbf{v}^1_{c_1} \concat\cdots \concat \mathbf{v}^1_{c_{|\mathcal{C}|}}
  \right]^\mathsf{T},
\end{align}
where
 $[\,\cdot\,;\, \,\cdot\,]$ represents the concatenation of row vectors.

% The Betti numbers are concatenated across different homology dimensions and cell types
%  to construct a fixed-dimensional feature vector for subsequent analysis.
% Let $\mathcal{C}$ be the set of all cell types, and 
%  let $\beta^{k}_{c}(\epsilon_j)$ denote the Betti curve 
%  calculated for cell type $c \in \mathcal{C}$ at homology dimension $k$. 
% For each filtration value $\epsilon_j$,
%  the Betti vector is constructed as
% \begin{align}
%   \mathbf{v}(\epsilon_i) = \left[ \beta^{0}_{c}(\epsilon_i), \, \beta^{1}_{c}(\epsilon_i) \right]_{c \in \mathcal{C}}.
% \end{align}
% Finally, by concatenating $\mathbf{v}(\epsilon_j)$ across all filtration values $\{\epsilon_j\}_{j=0}^{N_E}$, we obtain the full Betti vector
% \begin{align}
%   \mathbf{v} = \left[ 
%     \mathbf{v}(\epsilon_0); \, \mathbf{v}(\epsilon_1); \, \cdots ; \, \mathbf{v}(\epsilon_{N_E})
%   \right]^\mathsf{T},
% \end{align}
% where
%   $[\,\cdot\,;\, \,\cdot\,]$ represents the concatenation of row vectors.

This full Betti vector can be used
 as the input feature for downstream statistical and machine learning analyses.
In this study, 
 we use the Betti vector as the input of the proposed inverse surrogate model,
 where $N_{\mathrm{E}} = 1000$, $\epsilon_0 = 0$, and $\epsilon_i - \epsilon_{i-1} = \qty{0.1}{\micro m}$
 for all $i \in \{1,\ldots,N_\mathrm{E}\}$.
 \replaced{
This step size was chosen to balance biological resolution and model input dimensionality.
}
{
This step size was chosen based on the analysis shown in Appendix~\ref{ap:betti_step}.
}
 %the filtration values $\boldsymbol{\epsilon}$ were sampled uniformly with an interval of $0.1$ between consecutive elements.
To compute the persistent homology,
 we used the \texttt{ripser} Python library (version 0.6.12)~\cite{ctralie2018ripser}.

\subsection{Proposed TDA-based Inverse Surrogate Methods}
We use a GLM and an MLP
to estimate the simulator parameters $\boldsymbol{\Theta}$ from the full \replaced{Vetti}{Betti} vector $\mathbf{v}$ representing the observed biological pattern.
Let $D_{\Theta} = |\mathcal{C}|^2\times D_{\theta}$ denote the total number of the simulator's parameters,
 and the GLM, a mapping function $f_\mathrm{GLM}$ from $\mathbf{v}$ to $\boldsymbol{\Theta}$, is formulated as follows:
\begin{align}
% \boldsymbol{\theta}^{cc'} = \sigma\left(\mathbf{W}^{cc'} \mathbf{v}\right),
\boldsymbol{\Theta} = f_\mathrm{GLM}(\mathbf{v}) \coloneqq \sigma_0\left(\mathbf{W} \mathbf{v}\right),
\end{align}
where
  % $\theta_i\in\boldsymbol{\Theta}$ is a $i$-th simulator's parameter,
  $\mathbf{W}\!\in\!\mathbb{R}^{D_{\Theta}\times D_v}$ is a matrix of partial regression coefficients
  corresponding to $\boldsymbol{\Theta}$, and
  $\sigma_0(\cdot): \mathbb{R}^{D_\Theta}\rightarrow\mathbb{R}^{D_\Theta}$ is an element-wise nonlinear function.
To capture more complex dependencies between the parameters $\boldsymbol{\Theta}$ and the full Betti vector $\mathbf{v}$, 
 we also use an MLP with $L$ hidden layers
 to define a mapping function $\boldsymbol{\Theta}=f_\mathrm{MLP}(\mathbf{v})$, as follows:
\begin{align}
\mathbf{h}^{(1)} &= \sigma_1\left(\mathbf{W}^{(1)} \mathbf{v} + \mathbf{b}^{(1)}\right), \\
\mathbf{h}^{(l)} &= \sigma_l\left(\mathbf{W}^{(l)} \mathbf{h}^{(l-1)} + \mathbf{b}^{(l)}\right) \quad \text{for } l = 2, \dots, L, \\
\boldsymbol{\Theta} &= \sigma_{out}\left(\mathbf{W}^{(out)} \mathbf{h}^{(L)} + \mathbf{b}^{(out)}\right),
\end{align}
where $\mathbf{h}^{(l)}$ denotes the hidden representation at layer $l$, 
$\mathbf{W}^{(l)}$ and $\mathbf{b}^{(l)}$ are the weight matrix and bias vector of the $l$-th layer, and 
$\mathbf{W}^{(out)}$ and $\mathbf{b}^{(out)}$ are the weight matrix and bias vector of the output layer. 
We use the hyperbolic tangent
 as the nonlinear activation function $\sigma_l~(l\in\{0,1,\ldots,L,out\})$  in the GLM and the MLP given by
\begin{align}
\sigma_l(x) =\tanh(x) = \frac{e^x - e^{-x}}{e^x + e^{-x}}.
\end{align}

\subsection{Conventional Methods}
% We consider two types of conventional methods for comparison with the proposed method: a statistical pooling-based method and a PointNet-based method.
\added{
Because no established baseline exists for the inverse parameter surrogate task of multicellular patterns considered in this study,
}
we use two existing methods for general point cloud tasks as baselines: a statistical pooling-based method and a PointNet-based method.
\subsubsection{Statistical pooling-based method}
% First, as a baseline, we use
The statistical pooling-based method extracts hand-crafted global features of the observed biological pattern by directly applying pooling operations to the raw coordinates $\mathbf{X}^{c}$.
For each cell type $c \in \mathcal{C}$, we compute the maximum, minimum, and average values of the coordinates in $\mathbf{X}^c$ for each dimension, resulting in a row vector $\mathbf{s}^c \in \mathbb{R}^6$:
\begin{align}
\mathbf{s}^c = \left[ \text{max}(\mathbf{X}^c); \text{min}(\mathbf{X}^c); \text{avg}(\mathbf{X}^c) \right],
\end{align}
where the operation results $\max(\mathbf{X}^c)$, $\min(\mathbf{X}^c)$, and $\mathrm{avg}(\mathbf{X}^c)$ are specifically defined as follows:
% applied element-wise. 
% These operations can be viewed as a global max-pooling (and min/average pooling) applied directly to the raw coordinates $\mathbf{X}^c$.
\begin{align}
  \max(\mathbf{X}^c) &= \left[\max_{i\in\{1,\ldots,N_c\}} x^c_{i,0}~,~ \max_{i\in\{1,\ldots,N_c\}} x^c_{i,1}\right],   \\
  \min(\mathbf{X}^c) &= \left[\min_{i\in\{1,\ldots,N_c\}} x^c_{i,0}~,~ \min_{i\in\{1,\ldots,N_c\}} x^c_{i,1}\right],   \\
  \mathrm{avg}(\mathbf{X}^c) &= \left(\frac{1}{N_c} \sum_{i=1}^{N_c} \mathbf{x}^c_i\right)^{\!\!\!\mathsf{T}}.
\end{align}
% These operations can be viewed as a global max-pooling (and min/average pooling) applied directly to the raw coordinates $\mathbf{X}^c$.
The parameter set $\boldsymbol{\Theta}$ %for a specific pair of cell types $(c, c')$
is then estimated by considering the hand-crafted global features of all cell types as follows:
\begin{align}
% \boldsymbol{\theta}^{cc'} = f\left([\mathbf{s}^{c_1}; \mathbf{s}^{c_2}; \cdots; \mathbf{s}^{c_{|\mathcal{C}|}}]^\mathsf{T}\right),
\boldsymbol{\Theta} = f_{*}\left(\mathbf{s}\right), ~~~ \mathbf{s}=[\mathbf{s}^{c_1}; \mathbf{s}^{c_2}; \cdots; \mathbf{s}^{c_{|\mathcal{C}|}}]^\mathsf{T},
\end{align}
% Here, $f$ denotes the estimation model (such as a GLM, MLP, or othere model) that maps input features to the simulation parameters, depending on the method.
% In this case, we employed the GLM and the MLP for $f$.
where $*\!\!\in\!\!\{\mathrm{PoolingGLM}, \mathrm{PoolingMLP}\}$ and
 $f_{*}$ denotes a function that converts an input feature vector $\mathbf{s}$ into the simulation parameters $\boldsymbol{\Theta}$. 
% In this paper, we use the GLM and the MLP as $f$. 
The functions $f_\mathrm{PoolingGLM}$ and $f_\mathrm{PoolingMLP}$ 
 adopt the same architectures as their respective counterparts 
 used in the TDA-based inverse surrogate methods described in the previous subsection,
% Their architectures are identical to 
%  those used in the TDA-based inverse surrogate methods described in the previous subsection,
 except that 
 the input layer has dimension $6\times|\mathcal{C}|$ and
 the input feature vector is replaced by $\mathbf{s}$ instead of the full Betti vector $\mathbf{v}$. 

\subsubsection{PointNet-based method}
% The PointNet-based methods are a deep learning architecture designed for processing unordered point cloud data~\cite{qi2017pointnet,qi2017pointnet++}.
The PointNet‑based methods 
 incorporate a neural network architecture 
 designed to process unordered point cloud data~\cite{qi2017pointnet,qi2017pointnet++}.
Unlike the statistical pooling-based methods, 
 the PointNet maps the cell position into a latent feature representation 
 using an MLP-based function $h\colon \mathbb{R}^2 \!\!\rightarrow\!\!\mathbb{R}^{D_h}$,
 and then applies a global symmetric function (i.e., max-pooling) to achieve permutation invariance:
\begin{align}
% \mathbf{m}^c = \text{max-pooling}\left(\{h(\mathbf{x}^c_i)\}_{i=1}^{N_c}\right).
\mathbf{m}^c\!=\!\left[
 \max_{i\in\{1,\ldots,N_c\}} h_1(\mathbf{x}_i^c),
 \ldots,
 \max_{i\in\{1,\ldots,N_c\}} h_{D_h}(\mathbf{x}_i^c)
\right]\!\in\!\mathbb{R}^{D_h},
\end{align}
where $h_k(\cdot) ~ (k\in\{1,\ldots, D_h\})$ denotes the $k$-th element of the output vector of $h$.
Similar to the statistical pooling-based method, 
 the parameter set $\boldsymbol{\Theta}$ % for each pair of cell types $(c, c')$ 
 is then estimated by an MLP-based mapping function $f_\mathrm{PointNet}$ as follows:
\begin{align}
% \boldsymbol{\theta}^{cc'} = f\left([\mathbf{m}^{c_1}; \mathbf{m}^{c_2}; \cdots; \mathbf{m}^{c_{|\mathcal{C}|}}]^\mathsf{T}\right),
\boldsymbol{\Theta} = f_\mathrm{PointNet}\left(\mathbf{m}\right), ~~~ \mathbf{m} = [\mathbf{m}^{c_1}; \mathbf{m}^{c_2}; \cdots; \mathbf{m}^{c_{|\mathcal{C}|}}]^\mathsf{T}.
\end{align}
% where $f_\mathrm{PointNet}$ denotes the MLP.

PointNet++ further extends this architecture 
 by hierarchically capturing local geometric structures 
 via $L_{\mathrm{SA}}$ set abstraction layers. 
Procedures in each set abstraction layer are as follows.
For each cell type $c \in \mathcal{C}$, 
 each $l$-th set abstraction (SA) layer 
 % and $l \in \{1,2,\dots, L_{\mathrm{SA}}\}$ th set abstraction layer, 
\replaced{, where $l \in \{1,2,\dots, L_{\mathrm{SA}}\}$, takes the input points $\mathbf{X}^c$ and
 extracts $n^c_l$ representative points (centroids)
}{%
 ($l \in \{1,2,\dots, L_{\mathrm{SA}}\}$)
 applies the farthest point sampling algorithm
 to extract $n^c_l$ representative points (centroids)
 $\mathbf{X}^c_l=\{\grave{\mathbf{x}}^c_{l,j}\}^{n^c_l}_{j=1}$
 from the centroids $\mathbf{X}^c_{l-1}$
 obtained by the previous SA layer,
 where the first layer extracts centroids from $\mathbf{X}^c_{0}=\mathbf{X}^c$.
 % takes the input point set $\mathbf{X}^c_{l-1}$ from the previous set abstraction layer and
 % extracts $n^c_l$ representative points (centroids) for the current layer
}%
 % $\{ \grave{\mathbf{x}}^c_{l,j}\}^{n^c_l}_{j=1}$ 
 % are selected from the input points $\mathbf{X}^c_l$ 
 % using the farthest point sampling algorithm.
 % from selected points in previous layer $\mathbf{X}^c_{l-1}$.
% Here, the input points of the first layer $\mathbf{X}^c_0$ is set to $\mathbf{X}^c$.
% \added{
% Note that the first set abstraction layer $l=1$ takes all points $\mathbf{X}^c_0 =\mathbf{X}^c$ of cell type $c$.
% The centroids selected at $l$-th layer are denoted by $\mathbf{X}^c_l=\{\grave{\mathbf{x}}^c_{l,j}\}^{n^c_l}_{j=1}$.
% }
\replaced{
For each centroid $\grave{\mathbf{x}}^c_{l,j}$,
  we construct a group 
  $\mathcal{N}^c_{l,j}=\{\mathbf{x}^c_{l,j,i}\}_{i=1}^{p_{l,j}}$
    containing up to $p_{l,j}$ nearest neighbors in 
  $\mathbf{X}^c_l$ that satisfy $\|\mathbf{x}^c_{l,j,i} - \grave{\mathbf{x}}^c_{l,j}\|_{2} < r^{c}_l$,
}{%
For each centroid $\grave{\mathbf{x}}^c_{l,j}$,
 we construct an index set %represented by an index set
\begin{align}
  \!\!\mathcal{N}^c_{l,j}
  \!=\!
  \left\{
    % \mathbf{x}^c_{l-1,k} \in \mathbf{X}^c_{l-1},
    k \in \left\{1,\ldots,n^c_{l-1}\right\}
    \ \middle|\
    \left\|
      \mathbf{x}^c_{l-1,k}
      -
      \grave{\mathbf{x}}^c_{l,j}
    \right\|_2
    \!<
    r^c_l
  \right\}\!,\!
\end{align}
 where $r^c_l$ is a predefined radius and
 $p_{l,j}$ is a predefined number of nearest neighbors of  $\grave{\mathbf{x}}^c_{l, j}$,
 and $\mathcal{N}^c_{l,j}$ contains at most $p_{l,j}$ indices of points in $\mathbf{X}^c_{l-1}$.
 % containing up to $p_{l,j}$ nearest neighbors in $\mathbf{X}^c_{l-1}$,
}%%% End \replaced
% and are closest to $\grave{\mathbf{x}}^c_{l,j}$,
% \begin{align}
%   &= \mathrm{kNN}\left(
%   \grave{\mathbf{x}}^c_{l,j}, \mathcal{G}^c_{l,j}
%   \right)
%   %= \left\{\mathbf{x}^c_{l,j,i}\right\}^{p_{l,j}}_{i=1},
% \end{align}
 % is a predefined number of nearest neighbors.
If the number of selected points is less than $p_{l,j}$,
\replaced{
 we pad the set with a zero vector until it reaches size $p_{l,j}$.
}{%
 we add pseudo-index $0$ to the set,
 and this index is used in Eq. \eqref{eq:local_feature}.
}%
The local feature $\mathbf{m}^c_{l,j} \in \mathbb{R}^{D^c_l}$ 
 for each centroid $\grave{\mathbf{x}}^c_{l,j}$
 is then computed by 
 aggregating the points in $\mathcal{N}^c_{l,j}$ using a PointNet as follows:
\replaced{
\begin{align}
  \mathbf{m}^c_{l,j} 
  & = \left[ 
    \max_{i {\in} \left\{1,\ldots,p_{l,j}\right\}}
    \!\!m^c_{l,j,i,1}
    ,\ldots,
    \max_{i {\in} \left\{1,\ldots,p_{l,j}\right\}}
    \!\!m^c_{l,j,i,D^c_l}
    \right]\!\!,
    % \in\mathbb{R}^{D^c_l},
\\
  \mathbf{m}^c_{l,j,i}
  &= \left\{m^c_{l,j,i,d}\right\}_{d=1}^{D^c_l}   \nonumber   \\
  &=
    \gamma^c_{l}
    \left(\left[
        \mathbf{x}^c_{l,j,i}{-}\grave{\mathbf{x}}^c_{l,j}
        ; \mathbf{m}^c_{l-1,1}
        ; \cdots
        ; \mathbf{m}^c_{l-1,n^c_{l-1}}\right]\right),
\end{align}
}
{
\begin{align}
  \mathbf{m}^c_{l,j} 
  & = \left[ 
    \max_{k {\in} \mathcal{N}^c_{l,j}}
    m^c_{l,j,k,1}
    ,\ldots,
    \max_{k {\in} \mathcal{N}^c_{l,j}}
    m^c_{l,j,k,D^c_l}
    \right]\!\!,
    % \in\mathbb{R}^{D^c_l},
\\
  \mathbf{m}^c_{l,j,k}
  &= \left\{m^c_{l,j,k,d}\right\}_{d=1}^{D^c_l}   \nonumber   \\
  &= \begin{cases}
    \gamma^c_{l}\!
    \left(\left[
        \mathbf{x}^c_{l-1,k}{-}\grave{\mathbf{x}}^c_{l,j}
        ; \mathbf{m}^c_{l-1,k}
        \right]\right)\! \label{eq:local_feature} & (k\neq0)
    \\
    \gamma^c_{l}\!\left(\mathbf{0}_{2+D^c_{l-1}}\right) & (k=0)
    \end{cases},
\end{align}
}%
where 
\replaced{
 $\gamma^c_l\colon\mathbb{R}^{2+(n^c_{l-1}\times D^c_{l-1})}\!\rightarrow\!\mathbb{R}^{D^c_l}$ 
 }
 {
 $\gamma^c_l\colon\mathbb{R}^{2+ D^c_{l-1}}\!\rightarrow\!\mathbb{R}^{D^c_l}$ 
 }
 is an MLP that takes as input the concatenation of the relative position of the point 
 \replaced{
 $\mathbf{x}^c_{l,j,i}$
 }
 {
 $\mathbf{x}^c_{l-1,k}$
 }
to the centroid $\grave{\mathbf{x}}^c_{l,j}$ and
\replaced{
 the local feature $\mathbf{m}^c_{l-1,j}$ is an output of the previous layer,
 and  $\mathbf{m}^c_{0,j} = \emptyset$.
 }
 {
 the previous local feature $\mathbf{m}^c_{l-1,k}$ associated at that point,
 $\mathbf{0}_{2+D^c_{l-1}}$ denotes a (${2+ D^c_{l-1}}$)-dimensional zero vector,
 and $\mathbf{m}^c_{0,j}=\emptyset$.
 }
% of the point in the previous layer, which is $\emptyset$ at the first layer.
At the final set abstract layer $l=L_{\mathrm{SA}}$, 
 the local features 
 $\left\{\mathbf{m}^c_{L_{\mathrm{SA}},j}\right\}_{j=1}^{n_{L_\mathrm{SA}}^c}$ 
 \replaced{
 are concatenated as follows:
 \begin{align}
  \mathbf{m}^c = \left[\mathbf{m}^c_{L_{\mathrm{SA}},1}; \mathbf{m}^c_{L_{\mathrm{SA}},2}; \cdots; \mathbf{m}^c_{L_{\mathrm{SA}},n^c_{L_{\mathrm{SA}}}}\right]^\mathsf{T}.
\end{align}
 }
{
are aggregated by max pooling as follows:
 \begin{align}
   \mathbf{m}^c 
  & \!=\! \left[ 
    \max_{j {\in} \left\{1,\ldots,n^c_{L_{\mathrm{SA}}}\right\}}
    \!\!\!m^c_{L_{\mathrm{SA}},j,1}
    ,\ldots,
    \!\!\!\max_{j {\in} \left\{1,\ldots,n^c_{L_{\mathrm{SA}}}\right\}}
    \!\!\!m^c_{L_{\mathrm{SA}},j,D^c_{L_\mathrm{SA}}}
    \right]\!\!.
\end{align}
}%
The parameter set $\boldsymbol{\Theta}$ is then estimated by an MLP-based mapping function $f_\mathrm{PointNet}$ as follows:
\begin{align}
  \boldsymbol{\theta}^{cc'} \!\!\!=\! f_\mathrm{PointNet++}\!\left(\mathbf{m}\right), ~~ \mathbf{m} \!=\! [\mathbf{m}^{c_1}; \mathbf{m}^{c_2}; \cdots; \mathbf{m}^{c_{|\mathcal{C}|}}]^\mathsf{T}.
\end{align}

\subsection{Optimization}

\begin{comment}
Optimization:
- clocker:
- scaled_clockers: 5000 x 2000 (0 ~ 1に正規化されている)。
- simulartion parameters: 5000(5 * 1000) x 16

- X = scaled_clockers
    - X_train: 3500 / 5000: 3500 x 2000
- y = simulation_parameters
    - y_train: 3500 / 5000: 3500 x 16

- 学習の直前にXの2000次元の特徴ベクトルが16個飛ばしに間引かれている点が謎．
- W: 2000 x 16
    - initialize: np.random.rand

- Why is 5 replicates made for each sample?
  - Because the ABM-based simulation include randomness 
\end{comment}

We prepare a dataset of $N$ samples, 
 where each sample consists of 
 the cell positions $\mathbf{X}_n$ 
 and the corresponding ground-truth simulator parameters 
 $\hat{\boldsymbol{\Theta}}_n\in\mathbb{R}^{D_\Theta}_{>0}$
 given by
\begin{align}
\hat{\boldsymbol{\Theta}}_n
=\left\{\hat{\theta}^{cc'}_{nd}\right\}_{c,c'\in\mathcal{C},d\in\{1,\ldots,D_\theta\}}.
\end{align}
The cell positions $\mathbf{X}_n$ are converted 
 to the input feature vector $\mathbf{a}_n\!\in\!\{\mathbf{v}_n, \mathbf{s}_n, \mathbf{m}_n\}$
 corresponding to the function $f_*$, where $*\!\in\!\{\mathrm{GLM},\mathrm{MLP},\mathrm{PoolingGLM},\mathrm{PoolingMLP}, \mathrm{PointNet}\}$,
%of the proposed and conventional methods.
 and then we optimize the model parameters of $f_*$ by minimizing the sum of squared errors:
\begin{align}
  \mathcal{L}_\mathrm{SSE}
  &= \frac{1}{N}
    \sum_{n=1}^{N}
    % \sum_{c,c'\in\mathcal{C}}
    \left\|\hat{\boldsymbol{\Theta}}_n - f(\mathbf{a}_n)\right\|^2_2, 
  %   \\
  % \hat{\boldsymbol{\Theta}}_n
  % & = \left\{\hat{\theta}^{cc'}_{nd}\right\}_{c,c'\in\mathcal{C},d\in\{1,\ldots,D_\theta\}},
\end{align}
where
 $\|\cdot\|_2$ is the $L^2$ norm.
 % and $\hat{\boldsymbol{\Theta}}_n\!\in\!\mathbb{R}^{D_\Theta}_{>0}$
 % is a set of ground-truth simulator's parameters 
 % corresponding to the $n$-th set of cell positions $\mathbf{X}_n$.
 We used the adam optimizer for all methods.

\section{Experiments}

\subsection{Data} \label{subsec:data}

\begin{comment}
我々はXとYのペアデータを，エージェントベースモデルを用いたシミュレーションによって、提案手法の学習用に700個、提案手法のテスト用に300個作成した。

ABMはランダム性を含むため、1つのパラメータに対して5つのセル位置パターンを生成し、これら5つのパターンから計算されるベティベクトルの平均をそのパラメータに対応するペアとして使用する。

Because ABM involves randomness, five cell position patterns are generated for one parameter, and the average of the Betty vectors computed from these five patterns is used as the pair corresponding to that parameter.

\end{comment}

%%%% Please check if $\boldsymbol{\Theta}$ is correct -- this symbol only appears in the following paragraph. (kk)

We create $3721$ pairs of $\boldsymbol{\Theta}$ and $\mathbf{v}$ for training
 and $931$ pairs for testing our inverse surrogate model
 using the simulation with an ABM.
Since the simulation includes stochastic behavior,
we generate five variations of the cell position pattern $\mathbf{X}$
from a single parameter set $\boldsymbol{\Theta}$.
For proposed methods, the average of the Betti vectors calculated from these five patterns
is then used as the representative feature for that parameter set.
\replaced{
For conventional methods, we directory use the each input feature without averaging it across the five variations.
}
{
For the baseline methods, we directly use each input feature without averaging it across the five variations.
}
In addition,
 to investigate the performance of the inverse surrogate model,
 we manually craft examples of cell positions
 based on known variations in zebrafish pigment patterns,
 including \textit{dali/+} and \textit{leopard}.

% \textcolor{red}{
% We construct 
%  the training and evaluation dataset
%  $\{(\mathbf{v}_n, \hat{\boldsymbol{\theta}}_n^{cc'})\}_{n=1}^{N}$
%  for proposed inverse surrogate model
%  either synthetically or manually.
% For simplicity,
%  the subscript $n$ is omitted in the following.
% - The number of training data is 700 and the number of test data is 300, each of which is the averaged one of 5 replicates.
% }

\subsubsection{Agent-Based Model for Zebrafish Stripe}

% {\color{blue}
% \begin{itemize}
%     \item How did you define $N_\mathrm{diff}$?
%     \begin{itemize}
%         \item Your source code appears to set a limit on the total number of black and yellow cells $(=500)$.
%     \end{itemize}
%     % \item How to sample the candidate positions: ?.
%     \item Is the random birth of melanophores and xanthophores simultaneous or sequential? If simultaneous, they might appear in the same location. In your source code, if the conditions for black cell birth are not met at a location, the conditions for yellow cell birth appear to be checked at the same position. However, the original paper seems to indicate that the same number of candidate positions should be sampled for black and yellow cells.
%   \end{itemize}
% \end{itemize}
% }

\newcommand{\vxc}[0]{\mathbf{x}^{c}}
\newcommand{\ccount}[3]{\#\!\Big(\Omega_{\mathrm{#1}} ( \mathbf{x}^\mathrm{#2}_{i}(t) ), \mathbf{X}^\mathrm{#3}(t) \Big)}
\newcommand{\bcount}[2]{\#\!\Big(\Omega_{\mathrm{#1}} ( \overline{\mathbf{x}} ), \mathbf{X}^\mathrm{#2}(t) \Big)}

\begin{figure}[t]
\centerline{\includegraphics[width=\columnwidth]{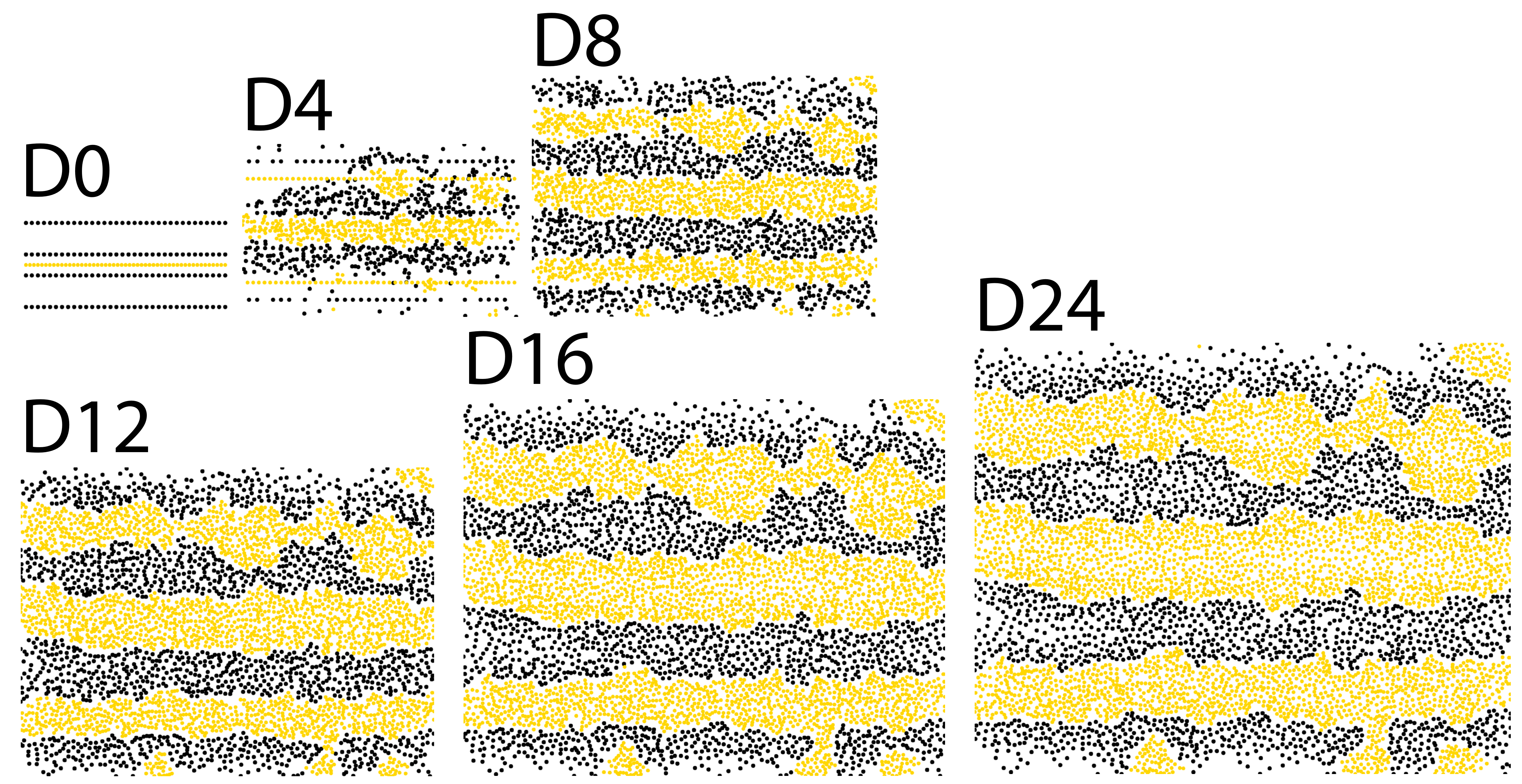}}
\caption{An example of ABM simulation results. All parameters for the ABM simulation were taken from~\cite{volkening_modelling_2015}.
 }
\label{fig_grow}
\end{figure}

\begin{comment}
The initial state (D0) shows four narrow black stripes and one yellow stripe. By Day 4 (D4), two additional narrow yellow stripes have appeared. By Day 24 (D24), the cell alignment forms a stripe pattern consisting of three black and three yellow lines. This simulation process reproduces wild-type zebrafish pigment pattern development. 
\end{comment}

An overview of the simulation process for zebrafish cell positions with ABMs
 is shown in Algorithm~\ref{alg:simulation} and Fig.~\ref{fig_grow}.
The set of cell positions $\mathbf{X}=\{\mathbf{X}^c\}_{c\in\mathcal{C}}$
 is derived from an ABM simulation using $\boldsymbol{\Theta}$
 generated by Latin Hypercube sampling.
The simulation variable $t\in [0, T]$, representing real-world time in days, is introduced
 for cell position variables: $\mathbf{X}^c(t)$ and $\mathbf{x}^c_i(t)$,
 where $\mathbf{X}^c(t)$ is a set of cell positions of type $c$ at time $t$,
 and $\mathbf{x}^c_i(t)$ is an $i$-th cell position of type $c$ at time $t$.
We also set $\mathcal{C} = \{\mathrm{B}, \mathrm{Y}\}$, 
 where $\mathrm{B}$ and $\mathrm{Y}$ represent the two types of iridophore relevant for the formation of zebrafish stripes: black melanophores and yellow xanthophores. %, respectively.

We initialize the set of cell positions $\mathbf{X}(0)=\{\mathbf{X}^c(0)\}_{c\in\mathcal{C}}$
 by evenly arranging cells in horizontal lines
 within a simulation domain of width $w_0$ and height $h_0$ as previously described in \cite{volkening_modelling_2015}.
% \textcolor{red}{- Initial locations}
Each black cell line contains $N_{\mathrm{B}}$ cells and is positioned
 $h \in \mathcal{H}^\mathrm{B}$~\unit{\micro m}
 from the top edge of the simulation domain,
 and each yellow cell line contains $N_{\mathrm{Y}}$ cells and
 is positioned $h\in \mathcal{H}^\mathrm{Y}~\unit{\micro m}$
 from the same edge.
The lines begin $l_{\mathrm{pad}}~\unit{\micro m}$ from the left edge and
 end $l_{{\mathrm{pad}}}~\unit{\micro m}$ from the right edge. %of the simulation domain.

%%%%%%%%%%% In the above paragraph (and others), I don't think \unit{\micro m} is necessary. (kk)

Between each time step, i.e., $t \in \mathbb{N}\cup\{0\}$,
 the simulation domain is stretched horizontally and vertically 
 by $k~\unit{\micro m}$ to recapitulate the growth of the developing zebrafish.
In proportion to the stretch of the simulation domain,
 the cells are also rearranged as follows:
\begin{align}
  \mathbf{x}^c_i(t) \leftarrow \mathbf{x}^c_i(t) \odot 
  \begin{pmatrix}
    1 + \frac{k}{w_t}    \\
    1 + \frac{k}{h_t}
  \end{pmatrix},
  \label{eq:rearrange}
\end{align}
where
 $\odot$ represents the Hadamard product,
 $w_t = kt+w_0$ and $h_t=kt+h_0$ are functions that return the width and height of the simulation domain at time $t$.
 %, and
 %$w_0$ and $h_0$ are the initial width and height of the simulation domain.

Furthermore, between each timestep, we simulate cell birth and death as a probabilistic event dependent on the neighborhood of each cell location.
Cell birth and death increase and decrease the elements of $\mathbf{X}^c(t)$ respectively.
To simplify notation,
 we define the function that counts points within a region on a two-dimensional plane as follows:
\begin{align}
\#(\mathcal{S}, \mathbf{X}) = \sum_{\mathbf{x}\in\mathbf{X}}\mathbf{1}_{\mathcal{S}}(\mathbf{x}),
\end{align}
where $\mathcal{S}$ is a region on a two-dimensional plane,
 $\mathbf{X}$ is a set of two-dimensional vectors representing the cell positions,
 and $\mathbf{1}_{\mathcal{S}}(\mathbf{x}):\mathbb{R}^2\rightarrow\{1,0\}$ is an indicator function defined by
\begin{align}
 \mathbf{1}_\mathcal{S}(\mathbf{x}) = \begin{cases}
    1 & \mathbf{x} \in \mathcal{S},  \\
    0 & \mathbf{x} \notin \mathcal{S}.
 \end{cases}
\end{align}

Let
 $\Omega_\mathrm{loc}(\mathbf{x}) \!=\!\left\{\mathbf{x}'\!\in\!\mathbb{R}^2\mid \|\mathbf{x}'-\mathbf{x}\|_2 \!\leq\! l_\mathrm{loc}\right\}$
 be the disk of radius $l_\mathrm{loc}$ centered at position $\mathbf{x}$ in a two-dimensional plane
 and
 $\Omega_\mathrm{podia}(\mathbf{x}) {=} \left\{\mathbf{x}'{\in}\mathbb{R}^2\mid l_\mathrm{podia} \leq \|\mathbf{x}'-\mathbf{x}\|_2 \leq l_\mathrm{podia} + l_\mathrm{width}\right\}$ 
 be the annulus of inner radius $l_\mathrm{podia}$ and width $l_\mathrm{width}$
 centered at position $\mathbf{x}$ in it,
 the rules for cell death are given as
\begin{align}
  \ccount{loc}{B}{Y} & > \mu \cdot \ccount{loc}{B}{B} \nonumber \\ 
  & \hspace{-12mm} \Rightarrow \text{death of $i$-th cell of type B}, \label{eq:death1} \\
  \ccount{loc}{Y}{B} & > \nu \cdot \ccount{loc}{Y}{Y} \nonumber \\ 
  & \hspace{-12mm} \Rightarrow \text{death of $i$-th cell of type Y}, \label{eq:death2} \\
  \ccount{podia}{B}{B} & > \xi \cdot \ccount{podia}{B}{Y} \nonumber \\ 
  & \hspace{-12mm} \Rightarrow
  \begin{minipage}{4.5cm}
    death of $i$-th cell of type B\\with probability $p_\mathrm{death}$ per day, \label{eq:death3}
  \end{minipage}
\end{align}
where $\mu$, $\nu$, and $\xi$ are hyperparameters obtained from \cite{volkening_modelling_2015}.

Let $\Omega_\mathrm{crowd}(\mathbf{x}) = \left\{\mathbf{x}'\in\mathbb{R}^2\mid \|\mathbf{x}'-\mathbf{x}\|_2\leq l_\mathrm{crowd}\right\}$
 be the disk of radius $l_\mathrm{crowd}$ centered at $\mathbf{x}$ in the two-dimensional plane,
 the rules for cell birth are given as
\begin{align}
  & \bcount{loc}{B} > \alpha \cdot \bcount{loc}{Y}, \nonumber \\ 
  & \bcount{loc}{Y} > \beta  \cdot \bcount{loc}{B}, \text{ and} \nonumber \\ 
  & \bcount{crowd}{Y} + \bcount{crowd}{B} < \eta \nonumber \\
  & \hspace{33mm} \Rightarrow \text{birth of a black cell at $\overline{\mathbf{x}}$}, ~\label{eq:birth_black}\\
% \end{align}
% \begin{align}
  & \bcount{loc}{Y} > \phi \cdot \bcount{loc}{B}, \nonumber \\ 
  & \bcount{loc}{B} > \psi  \cdot \bcount{loc}{Y}, \text{ and} \nonumber \\ 
  & \bcount{crowd}{Y} + \bcount{crowd}{B} < \kappa \nonumber \\
  & \hspace{30mm} \Rightarrow \text{birth of a yellow cell at $\overline{\mathbf{x}}$},~\label{eq:birth_yellow}
\end{align}
where
 $\alpha$, $\beta$, $\eta$, $\phi$, $\psi$, and $\kappa$
 are hyperparameters whose values are also obtained from \cite{volkening_modelling_2015}.
Cell birth can also be a stochastic event, and 
a melanophore or xanthophore can arise at a candidate location without sufficient neighboring cells with a probability $p^\mathrm{B}$ or $p^\mathrm{Y}$, respectively.

The cell positions are updated between each timestep
 by using the Euler method as follows:
\begin{align}
  \mathbf{x}^c_{i}(t+\delta t) \approx \vxc_{i}(t) + \frac{d}{dt}\vxc_{i}(t)\cdot \delta t,~\label{eq:move}
\end{align}
where
 $\delta t$ is a small time step in days, and
 the ordinary differential equation $d\mathbf{x}^c_{i}(t)/dt$ is defined by
\begin{align}
  \frac{d}{dt}\vxc_{i}(t) =
  &-\!\!\!\!\!\!\!\!\!\!
   \sum_{\mathbf{x}'\in\mathbf{X}^{c}(t)\setminus\{\mathbf{x}^c_i\}}
   \!\!\!\!\!\!\!\!\! \nabla Q^{cc}\Big(\mathbf{x}'-\vxc_{i}(t)\Big)  \nonumber \\
  &\ \ \ \ \ 
  -\!\!\!\!\!
   \sum_{c'\in \mathcal{C}\setminus\{c\}}
   \sum_{\mathbf{x}'\in\mathbf{X}^{c'}\!(t)} \!\!\! \nabla Q^{c'\!c}\Big(\mathbf{x}'-\vxc_{i}(t)\Big).
\end{align}
Here, $Q^{cc'}(\cdot)$ is the Morse potential function given by
\begin{align}
  Q^{cc'}(\mathbf{x})
  = R^{cc'}\exp\left\{-\frac{\|\mathbf{x}\|_2}{r^{cc'}}\right\}
  - A^{cc'}\exp\left\{-\frac{\|\mathbf{x}\|_2}{a^{cc'}}\right\},
\end{align}
where
 $\|\cdot\|_2$ is the $L^2$ norm,
 and the set of parameters $\boldsymbol{\theta}^{cc'}=\{R^{cc'},r^{cc'},A^{cc'},a^{cc'}\}$,
 i.e., $D_\theta = 4$,
 defines the behavior of interactions between the cells of type $c$ and type $c'$.
The parameters $R^{cc'}$ and $A^{cc'}$
 correspond to the strength scale of repulsion and attraction,
 and the parameters $r^{cc'}$ and $a^{cc'}$
 correspond to the length scale of repulsion and attraction.
To prevent the cells from moving outside the simulation domain,
 we adopt a reflective boundary condition as follows:
\begin{align}
  \vxc_{i}(t+\delta t) \leftarrow 
  f_\mathrm{ref}\left(\vxc_{i}(t+\delta t), \ \ 
  \begin{pmatrix}
      w_{\lfloor t \rfloor} & h_{\lfloor t \rfloor}\\
  \end{pmatrix}^\mathsf{T}\right),~\label{eq:bound}
\end{align}
where
 $\lfloor\cdot\rfloor: \mathbb{R}\rightarrow\mathbb{R}$ is the floor function and
 $f_\mathrm{ref}$ is an element-wise function defined by
\begin{align}
    f_{\mathrm{ref}}(x, L) = L - \left|x\ \mathrm{mod}(2 \times L) - L\right|,
\end{align}
where
 $\mathrm{mod}$ is the modulo operation.
 
As previously described in \cite{volkening_modelling_2015}, on the fourth simulated time step, we add a horizontal, single-layer stripe of $N_{\mathrm{add}}$ xanthophores from 20 and 80\% from the top of the domain to ensure consistent formation of three stripes when using experimentally determined parameters.
 % and $l_\mathrm{pad}~\unit{\micro m}$ is 
 % the padding size from the left and right sides of the simulation domain
 % to the lines containing the newly added cells.

\begin{comment}
- t<- t+dtを繰り返す．

シミュレーション領域の上端から２０％の位置において，左端からｘｍｍの位置から右からｘｍｍの位置まで，N個の黄色い細胞を等間隔に誕生させる．

と下端から２０％の位置にそれぞれ，
左端からｘｍｍの位置から右からｘｍｍの位置まで，N個の黄色い細胞を等間隔に誕生させる．
            
            x = np.linspace(50, self.domain.shape[1]-50, num_cells)
            y = np.linspace(position, position, num_cells)

                self.add_stripe_pattern_cells(0.2 * self.domain.shape[0], 48, 1)
                self.add_stripe_pattern_cells(0.8 * self.domain.shape[0], 48, 1)
                
\end{comment}

{
\setlength{\textfloatsep}{-5pt}
\SetAlgoSkip{}

\begin{algorithm}[t]
\caption{Simulation of cell positions with ABMs}\label{alg:simulation}
\KwIn{Initial cell positions $\mathbf{X(0)} = \{\mathbf{X}^c(0)\}_{c\in\mathcal{C}}$}
\KwOut{$\mathbf{X}(T) = \{\mathbf{X}^c(T)\}_{c\in\mathcal{C}}$}
\While{$t < T$}{
  Stretch the simulation domain and rearrange the cell positions based on~\eqref{eq:rearrange}\;
  Remove the cells based on~\eqref{eq:death1}, \eqref{eq:death2}, and \eqref{eq:death3}\;
  Add new cells based on Algorithm~\ref{alg:birth}\;
  $\tau \leftarrow t$\;
  \While{$t < \tau + 1$}{
    Move the cells based on ~\eqref{eq:move} and \eqref{eq:bound} \;
    $t \leftarrow t + \delta t$\;
  }
  \If{$t\ = 4$}{
    Add additional one-layer xanthophore horizontal stripes at 20 and 80\% locations\;
  }
}
\end{algorithm} 

\begin{algorithm}[t]
\caption{Simulation of cell birth}\label{alg:birth}
\KwData{Maximum number of cells to be born $N_{\mathrm{lim}}$
and that of attempts to simulate cell birth $N_{\mathrm{trial}}$}
$i\leftarrow 0$, $N_{\mathrm{birth}} \leftarrow 0$\;
\While{$i < N_\mathrm{trial}$}{
  \For{$j \leftarrow 1$ to $(N_\mathrm{lim} - N_\mathrm{birth})$}{
    Sample a candidate position $\overline{\mathbf{x}}\in\mathbb{R}^2$\;
    \If{there are no cells within $l_\mathrm{rand}~\unit{\micro m}$ of  $\overline{\mathbf{x}}$}{
      \If{$\overline{\mathbf{x}}$ satisfies the condition~\eqref{eq:birth_black}}{
        Add new black cell at position $\overline{\mathbf{x}}$\;
        $N_{\mathrm{birth}}\leftarrow N_{\mathrm{birth}} + 1$\;
      }\ElseIf{$\overline{\mathbf{x}}$ satisfies the condition~\eqref{eq:birth_yellow}}{
        Add new yellow cell at position $\overline{\mathbf{x}}$\;
        $N_{\mathrm{birth}}\leftarrow N_{\mathrm{birth}} + 1$\;
      }
    }\Else{
      Sample $q$ uniformly from $[0, 1]$\;
      \If{$q < p^\mathrm{B}$}{
        Add new black cell at position $\overline{\mathbf{x}}$\;
        $N_{\mathrm{birth}}\leftarrow N_{\mathrm{birth}} + 1$\;
      }\ElseIf{$q < p^{\mathrm{B} + \mathrm{Y}}$}{
        Add new yellow cell at position $\overline{\mathbf{x}}$\;
        $N_{\mathrm{birth}}\leftarrow N_{\mathrm{birth}} + 1$\;
      }
    }
  }
  $i\leftarrow i+1$\;
}
\end{algorithm}
}

\subsubsection{Manually Crafted Zebrafish Stripe}
To validate the effectiveness of the \TIIPS~framework, we generated patterns $\mathbf{X}$ as targets for parameter estimation. 
The generated patterns were based on known zebrafish mutant pigment patterns, including \textit{dali/+} and \textit{leopard}. 
Cell positions were manually placed within a domain with dimensions equal to those of the agent-based model. 
Cell types were specified during the manual placement, resulting in a total cell count of approximately 6000 cells 
(\textit{dali/+}: 5451 total cells, including 1179 melanophores and 4272 xanthophores; 
\textit{leopard}: 6204 total cells, including 1029 melanophores and 5175 xanthophores).
\begin{comment}
After all cell coordinate positions and respective cell types have been defined, we then specify a minimum distance threshold for each pair of cell types. This manually determined pattern serves as an initial configuration for testing the potential of our surrogate model in predicting the parameters on biologically relevant known target patterns.
\end{comment}
% \textcolor{red}{Question to Andrew - How did you create the zebrafish data with cell mutation? Please describe the algorithm for creating the data in detail.}

\subsection{Parameter Settings}
We set the values of the hyperparameters and variables for the ABM-based simulation
 according to~\cite{volkening_modelling_2015}.
The width and height of the simulation domain are initialized as
 $w_0 = \qty{2}{mm}$, and $h_0=\qty{1}{mm}$,
 and we extend the domain by $k=\qty{130}{\micro m}$ per day.
The parameters for the initial cell positions are set to
 $\mathcal{H}_{\mathrm{B}}=\{100, 400, 600, 900\}$
 $\mathcal{H}_{\mathrm{Y}}=\{500\}$,
 $N_\mathrm{B}=34$, and $N_\mathrm{Y}=51$.
The hyperparameters of the rules for cell death are set to $\mu=1$, $\nu=1$, and $\xi=1.2$, and
 those for cell birth are set to $\alpha=1$, $\beta=3.5$, $\eta=6$, $\phi=1.3$, $\psi=1.2$, and $\kappa=10$.
The radii and width defining the region of the disks or annulus are given as
 $l_\mathrm{loc} = \qty{75}{\micro m}$, $l_\mathrm{podia} = \qty{318}{\micro m}$,
 $l_\mathrm{width} = \qty{25}{\micro m}$, $l_\mathrm{crowd} = \qty{82}{\micro m}$,
 and $l_\mathrm{rand} = \qty{82}{\micro m}$.
The probabilities related to the cell death and birth
 are set to $p_\mathrm{death}=0.0333$, $p^\mathrm{B}=0.03$, and $p^\mathrm{Y}=0.005$.
For the simulation of cell birth,
 we set the maximum number of cells to be born $N_{\mathrm{lim}}$ to $500$
 and that of attempts to simulate cell birth $N_{\mathrm{trial}}$ to $2$.
The parameters $\{R^{cc'}, A^{cc'}\}_{c,c'\in\{\mathrm{B},\mathrm{Y}\}}$
 are sampled within the range of $0$ to $1000$,
 while the parameters $\{r^{cc'}, a^{cc'}\}_{c,c'\in\{\mathrm{B},\mathrm{Y}\}}$
 are sampled within the range of $1$ to $100$.
In the ABM simulation of cell development,
 these sampled values remain unchanged.
On the other hand,
 we use the normalized parameters from $0$ to $1$
 as the ground-truth simulator's parameters.
We set the simulation period to $T=24$
 and the small time step to $\delta t = 1$.

For all methods, input features were normalized to a mean of $0$ and a standard deviation of $1$, and output targets were scaled to a range between $-1$ and $1$. 
We used the Adam optimizer to minimize the mean squared error (MSE) as the loss function. Early stopping based on validation loss with a patience of $10$ epochs was employed for every experiment.
The proposed \TIIPS~GLM and its pooling-based counterpart were trained with a learning rate of $5.0 \times 10^{-6}$ and a batch size of $512$. 
For the MLP-based models (both \TIIPS~and pooling-based), we used three hidden layers with $1024$ hidden units each, a hyperbolic tangent ($\tanh$) activation, 
%\replaced
%{
%a learning rate of $1.0 \times 10^{-4}$, a batch size of $512$, a dropout rate of $0.2$, and a weight decay of $0.0001$.
%}
a learning rate of $5.0 \times 10^{-6}$, a batch size of $64$.
The PointNet method utilized a dual-branch architecture to process melanophore and xanthophore positions separately. 
Each branch consisted of an MLP $h$ with two linear layers and ReLU activation that mapped input features to a $D_h = 64$-dimensional embedding space. 
The  MLP $f_{\mathrm{PointNet}}$ consisted of three hidden layers with $128$ hidden units each, using $\tanh$ activation, a learning rate of $1.0 \times 10^{-3}$, a dropout rate of $0.1$, and a weight decay of $0.0001$.
For PointNet++, we also implemented a two-branch architecture. Each branch contained two set abstraction layers. 
The first layer used $64$ representative points, a neighborhood radius of $0.2$, and $16$ samples with MLP layers of $[32, 32, 64]$. 
The second layer used $16$ points, a radius of $0.4$, and $32$ samples with MLP layers of $[64, 128, 256]$. 
The concatenated outputs were processed through an MLP $f_{\mathrm{PointNet++}}$ with one fully connected layer containing $128$ hidden units and batch normalization. 
The optimization used a learning rate of $1.0 \times 10^{-5}$, a batch size of $512$, a dropout rate of $0.4$, and a weight decay of $0.001$.
\added{
For all models and training conditions, we repeated the experiments 5 times with different random seeds: 0, 1, 2, 3, and 4. 
The mean and standard deviation across these runs were used to assess variability due to random initialization, mini-batch sampling, and other stochastic factors in the training procedure.
}

\subsection{Evaluation Metrics}

 %\textcolor{red}{Question to Andrew -The explanation about the evaluation metrics is correct?}

We employed three metrics to evaluate the outcome of the inverse surrogate model.
The first is the mean squared errors (MSE) 
% for the inferred parameters by the surrogate model, which is 
defined by
% by the equation below.
\begin{align}
\mathrm{MSE} = \frac{1}{N}\sum_{n=1}^{N} \sum_{c,c' \in C} \left\| \hat{\boldsymbol{\theta}}^{cc'}_n-\boldsymbol{\theta}^{cc'}_n \right\|^2,
\end{align}
where $\hat{\boldsymbol{\theta}}^{cc'}_n$ is a set of true parameters.
The second metric is the Pearson correlation coefficient between the estimated parameters and the true parameters, which is defined by
\begin{align}
r^{cc'}_{d} = 
&\frac{\sum^{N}_{n=1} \left( \theta^{cc'}_{nd} - \bar{\theta}^{cc'}_{d} \right)\left( \hat{\theta}^{cc'}_{nd} - \bar{\hat{\theta}}^{cc'}_{d} \right) }
{
  \sqrt{
    \sum^{N}_{n=1} \left( \theta^{cc'}_{nd} - \bar{\theta}^{cc'}_{d} 
    \vphantom{\bar{\hat{\theta}}} \right)^2
  }
  \sqrt{
    \sum^{N}_{n=1} \left( \hat{\theta}^{cc'}_{nd} - \bar{\hat{\theta}}^{cc'}_{d} \right)^2
  }
},
\end{align}
where $\theta^{cc'}_{nd}$ and $\hat{\theta}^{cc'}_{nd}$ are 
 the $d$-th estimated and true parameters, respectively,
 for the $n$-th sample of the $c$ and $c'$ cell types,
 and $\bar{\theta}^{cc'}_{d}$ and $\bar{\hat{\theta}}^{cc'}_{d}$ are the mean values defined by
\begin{align}
 \bar{     \theta }^{cc'}_d = \frac{1}{N}\sum_{n=1}^{N}      \theta^{cc'}_{nd}, &&
 \bar{\hat{\theta}}^{cc'}_d = \frac{1}{N}\sum_{n=1}^{N} \hat{\theta}^{cc'}_{nd}.
\end{align}
 % of $\theta^{cc'}_{nd}$ and $\hat{\theta}^{cc'}_{nd}$ across all samples in the testing set.
We averaged the Pearson correlation coefficients across all parameters and all pairs of cell types to obtain a single correlation value $r$ for each method as follows:
\begin{align}
r = \frac{1}{
  D_{\Theta} % D_{\theta} |\mathcal{C}|^2
} \sum_{d=1}^{D_{\theta}} \sum_{c,c' \in \mathcal{C}} r_{d}^{cc'}.
\end{align}
The final metric is the ratio of the standard deviation of the estimated parameters to that of the true parameters, defined as follows:
\begin{align}
  \sigma_{\mathrm{ratio}} = \frac{1}{D_{\Theta} 
    % D_{\theta} |\mathcal{C}|^2
  } \sum_{d=1}^{D_{\theta}} \sum_{c,c' \in \mathcal{C}} 
  \sqrt{\frac{\sum^N_{n=1} \left( \theta^{cc'}_{nd} - \bar{\theta}^{cc'}_{d} \vphantom{\bar{\hat{\theta}}} \right)^2}{\sum^N_{n=1} \left( \hat{\theta}^{cc'}_{nd} - \bar{\hat{\theta}}^{cc'}_{d}  \right)^2}}.
\end{align}
This metric is also averaged across all parameters and all pairs of cell types to obtain a single value for each method.
Note that the first metric is computed using parameters normalized to the range $[-1, 1]$, where the normalization scales were determined based on the training set.

\subsection{Results}

\subsubsection{Validation with Simulated Data}

\begin{figure*}[t]
\centerline{\includegraphics[width=0.9\textwidth]{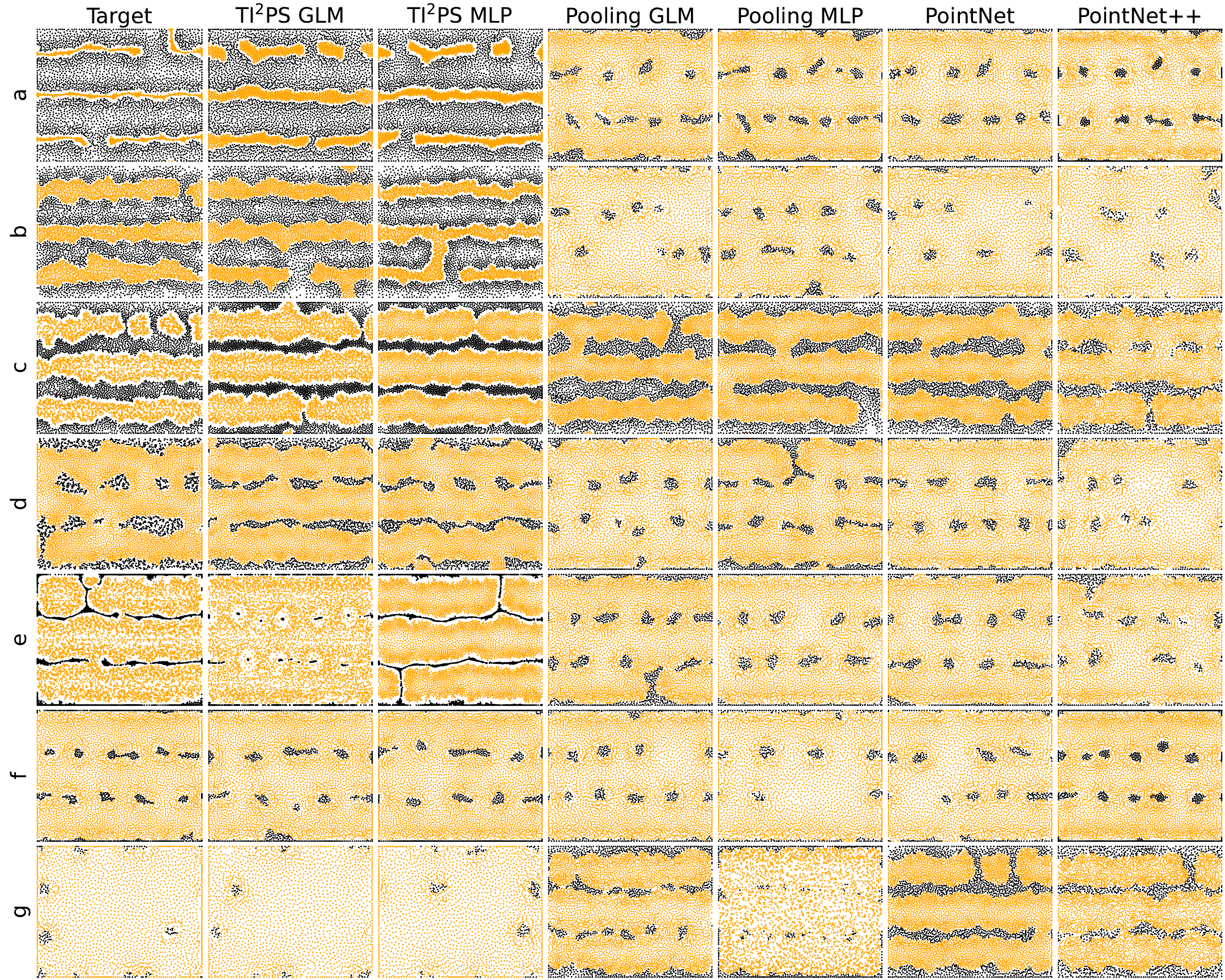}}
\caption{
Visual comparisons of simulated patterns using parameters estimated by each method. 
The leftmost column shows seven representative target patterns from the test set, 
including stripes (rows a--e) and dots (rows f and g). These targets exhibit variations in cell distribution, 
such as black-dominant (row a), yellow-dominant (rows d and e), and balanced (rows b and c) patterns. 
Columns 2--7 show simulation results based on the parameters estimated by the proposed and conventional methods. 
The second and third columns represent the proposed \TIIPS~GLM and \TIIPS~MLP, while the remaining columns show results from the conventional methods.
\added{
All simulated patterns, except for the target pattern, were generated using the corresponding models trained with random seed 0.
}
}
\label{fig_result_simulated}
\end{figure*}

The proposed methods successfully estimated the parameters
 that can reproduce spatial patterns similar to the target patterns. 
Fig.~\ref{fig_result_simulated} shows the visual comparison
 between the target patterns and the simulation results
 obtained using the proposed and conventional methods. 
\added{The examples shown in Fig.~\ref{fig_result_simulated} were selected as representative samples from the target set, rather than as successful cases for the proposed methods.}
The target patterns differ from one another
 in both large‑scale features, such as pattern types like stripes or polka dots, 
 and small‑scale features, such as the areas occupied by each cell type and the gaps between cells.

\replaced{
The proposed methods, shown in the second and third columns of Fig.~\ref{fig_result_simulated}, 
 successfully reproduced both large- and small-scale features.
In particular,
 \TIIPS~MLP reproduced fine details,
 such as the small gaps between different cell types in Fig.~\ref{fig_result_simulated}c 
 and the thin black stripes in Fig.~\ref{fig_result_simulated}e.
\TIIPS~GLM was unable to reproduce these fine details,
 suggesting that nonlinear models are better at capturing complex pattern features.
 }
 {
The proposed methods, shown in the second and third columns of Fig.~\ref{fig_result_simulated},
successfully reproduced major features of the target patterns.
Both \TIIPS~GLM and \TIIPS~MLP mostly reproduced large-scale features, such as stripes and polka dots.
However, local and global discrepancies persisted in some cases.
For \TIIPS~GLM, a global discrepancy was observed in Fig.~\ref{fig_result_simulated}e, where a polka-dot pattern appeared in the simulation although it was not present in the corresponding target pattern.
For \TIIPS~MLP, no similarly obvious global discrepancy was observed in these examples.
Nevertheless, local discrepancies remained, such as differences in the gap size between yellow and black cells in Fig.~\ref{fig_result_simulated}b and the density cells in Fig.~\ref{fig_result_simulated}c.
 }
% Unlike the proposed methods, 
%  the conventional methods often failed to reproduce the target patterns. 
% These methods frequently failed to capture even the large-scale features. 
Unlike the proposed method,
 the conventional methods often failed to reproduce the target patterns, even the large‑scale features.
For example, as shown in Fig.~\ref{fig_result_simulated}a, \ref{fig_result_simulated}b, \ref{fig_result_simulated}d, and \ref{fig_result_simulated}e, 
 they produced polka‑dot patterns even when the target patterns were stripes.
% For example, they produced polka-dot patterns instead of the target stripe patterns in Fig. \ref{fig_result_simulated}a, b, d, and e. 
They correctly reproduced only
 the stripe pattern in Fig. \ref{fig_result_simulated}c and 
 the polka-dot pattern in Fig. \ref{fig_result_simulated}f.
% Table \ref{tab_combined_evaluation}
%  summarizes the evaluation metrics for the estimated parameters.
Table~\ref{tab_combined_evaluation} summarizes
 the evaluation results of parameter estimates across different metrics.
The proposed \TIIPS~MLP method
 achieved the lowest MSE, the highest correlation $r$, and the best $\sigma_{\mathrm{ratio}}$. 
\replaced{
PointNet++
 achieved the second-lowest MSE %for parameter estimation
 despite failing to reproduce the visual patterns. 
However, its Pearson correlation was nearly zero. 
For this method,
 the low correlation and high standard deviation compared to other conventional methods 
 suggest that the parameter estimates are highly random.
 % the estimated parameters have high randomness.
The standard deviation ratio for the pooling GLM and MLP 
 was lower than for the other methods. 
 }
 {
The baseline methods generally exhibited higher MSE values and lower Pearson correlation coefficients than the proposed \TIIPS~MLP and GLM.
$\sigma_{\mathrm{ratio}}$ varied across the baseline methods, with PointNet showing the lowest value and PointNet++ the highest.
The differences in the mean values were substantially larger than the standard deviations across random seeds for all methods.
 }
These results show 
 that the \TIIPS~framework outperforms the conventional methods
 in both visual pattern reproduction and parameter estimation accuracy.

\begin{table}[t]
  \centering
  \caption{\replaced{Performance comparison: MSE, Pearson correlation ($r$), and standard deviation ratio ($\sigma_{\mathrm{ratio}}$).}{Performance comparison in terms of MSE, Pearson correlation ($r$), and standard deviation ratio ($\sigma_{\mathrm{ratio}}$). Values are reported as the mean $\pm$ standard deviation over five independent runs with different random seeds.}}
  \label{tab_combined_evaluation}
  \begin{tabular}{l | c c c}
    \toprule
    Method & \makecell{MSE ($\downarrow$)} & \makecell{Pearson $r$ ($\uparrow$)} & \makecell{$\sigma_{\mathrm{ratio}}$ (\%) ($\uparrow$)} \\
    \midrule
    \makecell[l]{TI$^{2}$PS GLM (Ours)}
    & \makecell{$0.2476$ \\ $\pm 0.0021$}
    & \makecell{$0.4933$ \\ $\pm 0.0068$}
    & \makecell{$53.80717$ \\ $\pm 0.4956$} \\

    \makecell[l]{TI$^{2}$PS MLP (Ours)}
    & \makecell{$\mathbf{0.1911}$ \\ $\mathbf{\pm 0.0016}$}
    & \makecell{$\mathbf{0.6360}$ \\ $\mathbf{\pm 0.0038}$}
    & \makecell{$\mathbf{68.5540}$ \\ $\mathbf{\pm 1.5241}$} \\

    \midrule
    Pooling GLM
    & \makecell{$0.3533$ \\ $\pm 0.0003$}
    & \makecell{$-0.0039$ \\ $\pm 0.0015$}
    & \makecell{$18.2502$ \\ $\pm 0.1324$} \\

    Pooling MLP
    & \makecell{$0.3677$ \\ $\pm 0.0009$}
    & \makecell{$-0.0008$ \\ $\pm 0.0005$}
    & \makecell{$27.7520$ \\ $\pm 0.4176$} \\

    PointNet
    & \makecell{$0.351$ \\ $\pm 0.0006$}
    & \makecell{$-0.0054$ \\ $\pm 0.0006$}
    & \makecell{$16.2239$ \\ $\pm 0.5604$} \\

    PointNet++
    & \makecell{$0.4203$ \\ $\pm 0.0048$}
    & \makecell{$0.0019$ \\ $\pm 0.0018$}
    & \makecell{$49.1806$ \\ $\pm 1.3653$} \\

    \bottomrule
  \end{tabular}
\end{table}

\subsubsection{Sensitivity to Training Data Size}

\begin{figure*}[t]
    \centering
    \includegraphics[width=0.9\textwidth]{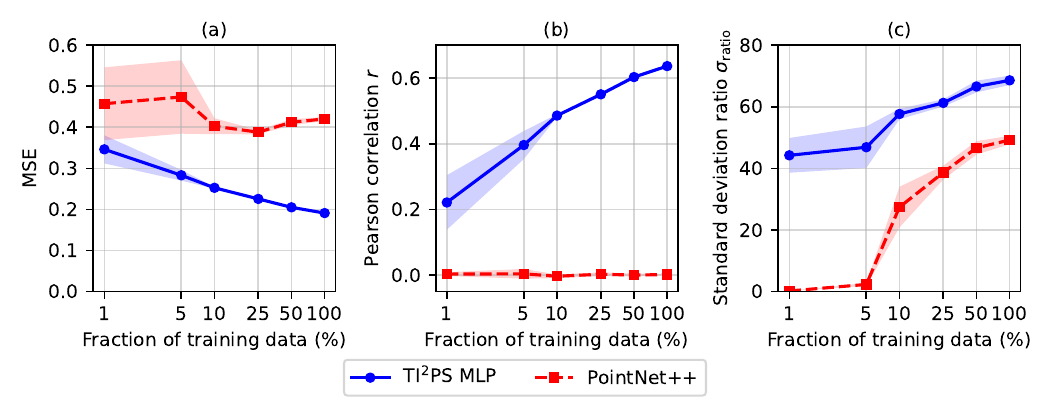}
    \caption{
    \replaced{
      Sensitivity analysis of the training data amount for the proposed \TIIPS~MLP and PointNet++ methods: (a) MSE, (b) Pearson correlation $r$, 
      and (c) standard deviation ratio $\sigma_{\mathrm{ratio}}$. 
      The proposed \TIIPS~MLP method shows superior performance and stability across all metrics, 
      even with limited training data. Missing data points in (a) PointNet++ with 10\% training data indicate that it fails to converge during training.
      }
      {
        Sensitivity analysis of the training data amount for the proposed \TIIPS~MLP and PointNet++ methods: 
        (a) MSE, (b) Pearson correlation $r$, and (c) standard deviation ratio $\sigma_{\mathrm{ratio}}$. 
        The proposed \TIIPS~MLP method shows superior performance and stability across all metrics, even with limited training data. 
        Markers indicate the mean values over five independent runs with different random seeds, and the shaded regions indicate the corresponding standard deviation ranges.
      }
    }
    \label{fig_data_reduction}
\end{figure*}

Fig.~\ref{fig_data_reduction}
 shows how the model performance changes with different training data sizes. 
In Fig.~\ref{fig_data_reduction}(a),
 the proposed TI$^2$PS~MLP method achieved a lower MSE than PointNet++ 
 in all data sizes.
\replaced{
Notably, 
 the proposed method, which uses only 10\% of the training data,
 achieved a lower MSE than PointNet++ trained on full dataset. 
 }
 {
 Notably,
Notably, the proposed method, despite being trained with only 10\% of the training data, outperformed PointNet++ trained on the full dataset across all evaluation metrics.
 }
\removedtext{
The MSE of PointNet++ is unstable with small datasets and 
 fails to converge with 10\% of the training data.
 }
Fig.~\ref{fig_data_reduction}(b)
 shows the correlation $r$. 
The correlation $r$ of the proposed method 
 increases steadily as the training data size increases.
In contrast, 
 that of PointNet++ remains near zero and shows almost no change.
Fig.~\ref{fig_data_reduction}(c)
 shows the standard deviation ratio $\sigma_{\mathrm{ratio}}$. 
For both methods, 
 this ratio increases as the training data size increases. 
The standard deviation ratio of PointNet++ 
 shows an increasing trend starting at 10\% of the training data.
On the other hand, 
 the standard deviation ratio of the proposed method is consistently higher than that of PointNet++.
These results indicate 
 that the proposed method is more effective at learning the parameter distributions
 across different amounts of training data.
 
\subsubsection{Validation with Manually Crafted Mutant Patterns}

\begin{figure}[t]
\centering
\includegraphics[width=0.9\columnwidth]{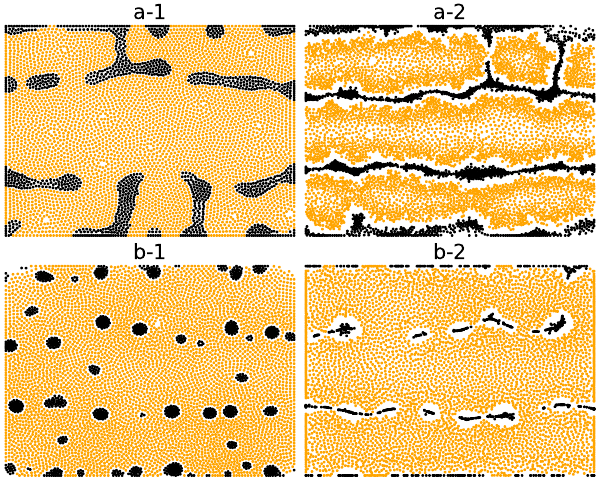}
\caption{
Validation result with manually crafted mutant data: 
(a-1) target distorted stripe pattern of \textit{dali/+} mutant and 
(a-2) simulated pattern using the parameters estimated from the target;
(b-1) target polka-dot-like pattern of \textit{leopard} mutant and 
(b-2) simulated pattern using the parameters estimated from the target. 
}
\label{fig_result_mutant}
\end{figure}

We further evaluated the \TIIPS~MLP framework using manually crafted mutant patterns.
Fig. \ref{fig_result_mutant} compares the crafted mutant patterns with the patterns generated by the ABM using the estimated parameters.
The simulated results shared several features with the crafted patterns. 
First, in both \textit{dali/+} and \textit{leopard}, the simulated patterns produced a dominant xanthophore population, 
which matches the crafted patterns (Fig. \ref{fig_result_mutant} a-1 and b-1). 
Second, the \textit{dali/+} simulation captured the incomplete formation of melanophore stripes seen in the crafted pattern (Fig. \ref{fig_result_mutant} a-2). 
Third, the \textit{leopard} simulation produced small melanophore clusters similar to those in the crafted pattern (Fig. \ref{fig_result_mutant} b-2).
However, there were also differences between the crafted and simulated patterns. 
For the \textit{dali/+} mutant, the simulation showed a larger gap between the melanophores and xanthophores than the crafted pattern. 
Similarly, for the \textit{leopard} mutant, the gap between different cell types was larger, and the size of the melanophore clusters was smaller than in the crafted pattern.
Table \ref{tab:parameter_comparison} summarizes the estimated parameters for the wild type and crafted mutant patterns with \TIIPS~MLP.
The estimated parameters for each mutant showed specific trends. 
Compared with the wild type, both mutant patterns showed higher repulsion scales for interactions within the same cell type ($R^{BB}$ and $R^{YY}$). 
The repulsion scale for interactions between different cell types ($R^{BY}$) was lower in the dali/+ pattern but higher in the leopard pattern than in the wild type. 
The attraction scale ($A^{BY}$) showed no major differences across the patterns. 
These values represent the intercellular interactions underlying each pattern. 
This suggests that the differences in the estimated parameters account for the cell-level phenotypes observed in each mutant.
%leopardは異種間の強い反発と多数xathophoreがmelanophoreの小さな集合を作る
%daliのmelanophore同士の中距離引力が低い、異種間相互作用のうちxanthophoreから受けるmelanophoreの異種間反発力が低いのでのmelanophoreの領域が広がる結果となった。

\begin{comment}
Although the surrogate-model predicted ABM is unable to recapitulate the exact behavior of the target mutant iridiphore patterns, we are able to identify some potential cell-cell interactions that may bring about the mutant behaviors. In both mutant patterns, the model predicts high, long-range repulsion for xanthophores ($R^{xx'}$, $r^{xx'}$) to produce a predominant xanthophore field in the simulation domain. In the \textit{leopard} case, the model predicts no melanophore repulsion ($R^{mm'}$) but high melanophore attraction ($A^{mm'}$) for low ranges ($a^{mm'}$), and high repulsion at high ranges for heterogeneous cell type interactions (($R^{mx'}$, $r^{mx'}$) and ($R^{xm'}$, $r^{xm'}$). This indicates that strong repulsion forces coupled with large numbers of xanthophores forces melanophores to form small spots instead of the wild-type stripes. In the \textit{dali/+} case, the surrogate model predicts low attraction at moderate ranges for melanophores ($A^{mm'}$,  $a^{mm'}$) and lower heterogeneous cell type repulsions (($R^{mx'}$, $r^{mx'}$) and ($R^{xm'}$, $r^{xm'}$). This indicates that a lower heterogeneous repulsion force allows the melanophores to form larger regions in the simulation domain than in the \textit{leopard} mutant, creating the observed incomplete melanophore stripes.
\end{comment}

\begin{table}[htbp]
\centering
\caption{Estimated parameters from wild type and mutant patterns.}
\label{tab:parameter_comparison}
\begin{tabular}{lccc}
\toprule
Parameter & Wild type (stripe) & \textit{dali/+} & \textit{leopard} \\
\midrule
$R^{BB}$ & 250.00 & 941.64 & 858.84 \\
$r^{BB}$ & 20.00  & 8.89   & 0.38   \\
\addlinespace
$R^{YY}$ & 200.00 & 938.39 & 901.65 \\
$r^{YY}$ & 11.00  & 8.88   & 77.44  \\
\addlinespace
$R^{BY}$ & 450.00 & 214.23 & 813.95 \\
$r^{BY}$ & 20.00  & 52.47  & 94.87  \\
\addlinespace
$R^{YB}$ & 550.00 & 777.77 & 723.39 \\
$r^{YB}$ & 20.00  & 72.07  & 65.44  \\
\addlinespace
$A^{BY}$ & 650.00 & 509.26 & 787.55 \\
$a_{BY}$ & 12.00  & 11.98  & 9.24   \\
\bottomrule
\end{tabular}
\end{table}

\begin{comment}
\begin{figure}[t]
\centerline{\includegraphics[width=\columnwidth]{fig/leopard_like_mutant_pattern.pdf}}
\caption{Validation result  \textit{leopard} pigment pattern (a) target manually crafted pigment pattern (b) simulated pigment pattern with estimated parameters}
\label{fig_result_leopard}
\end{figure}
\end{comment}

\section{Discussion and Conclusion}

%Paragraph 1: Summarise the overall procedures of this research.
%
This study proposed the TI$^{\mathbf{2}}$PS framework
 based on an inverse surrogate modeling approach
 for estimating parameters in ABMs from spatial cell patterns through a feature derived from TDA.
To robustly and effectively quantify the global structure of cell patterns,
 we employed the Betti vector, a topological descriptor derived from TDA.
% For robust quantification of cell patterns, 
%  we implemented the Betti vector, 
%  which is a topological descriptor derived from TDA that
%  effectively captures the global structure of cell patterns.
While we demonstrated the utility of the framework for pigment pattern formation in zebrafish,
 the framework is designed to be applicable to diverse biological systems governed by 
 multicellular interactions.
This framework provides a foundation for modeling and refining agent-based models of biological systems.

%Paragraph 2: key result and findings : The poroposed method was scceeded on test case.
% 1 The proposed method successfully estimated the parameters that visually reproduces the target patterns.
% -> indicates that a learnable mapping exists from ABM simulation results to their corresponding parameters
% -> this finding is important to following research on solving inverse problems of stochastic multicellular systems.
% 2 The proposed method visually and quantitively outperformed the conventional methods for general point cloud classification and segmentation.
% -> indicates that the fundamental differences of general targets of the point cloud classification with the problem of multicellular pattern inverse problem.
% -> The validation results of TDA-based methods and Pooling-based methods suggest that the Betti vector derived from TDA is a key of the success of the proposed method.
% 3 The proposed method is robust to limited training data, rather than the pointnet++, which is a state-of-the-art in the conventional methods.
% -> This feature is particularly important for biological application, where the amount of measurement data is often limited due to cost and ethical constraints.

The results clearly demonstrate that the \TIIPS~MLP model %significantly
 outperforms all baseline methods across all evaluated metrics (Table~\ref{tab_combined_evaluation}), 
achieving not only a lower MSE but also a higher Pearson correlation ($r$) and output variance ratio ($\sigma_{\mathrm{ratio}}$). 
This model successfully reproduces the precise visual features of the target patterns, 
ranging from global structures such as stripes (Figs.~\ref{fig_result_simulated}a–\ref{fig_result_simulated}e) and dots (Figs.~\ref{fig_result_simulated}f and \ref{fig_result_simulated}g).
\removedtext{to local features, 
such as the small gaps between different cell types within the same global pattern (Figs.~\ref{fig_result_simulated}b and \ref{fig_result_simulated}c).}
Such high performance indicates that
 the Betti vector effectively captures the multiscale topological features of cellular arrangements, 
 allowing the model to accurately map ABM simulation outputs to their underlying parameters. 
Furthermore, the fact that the MLP outperforms the GLM indicates 
 that the relationship between the Betti vector and the simulator parameters is intrinsically nonlinear. 
% This is particularly evident in the significant improvement in the Pearson correlation and the visual contrast between the GLM and MLP results (e.g., Fig. \ref{fig_result_simulated}e), 
% confirming that modeling this nonlinearity is essential for precise pattern recognition and parameter estimation.
The substantial improvement in the Pearson correlation and 
 the visual contrast between the GLM and MLP results (e.g., Fig.~\ref{fig_result_simulated}e) 
 also confirm 
 that modeling this nonlinearity is essential for accurate pattern recognition and parameter estimation.

% In contrast, 
The pooling-based GLM and MLP
 showed much lower correlation coefficients $r$ and variance ratios $\sigma_{\mathrm{ratio}}$. 
% As seen in Fig. \ref{fig_result_simulated}, they mostly predicted similar "polka-dot" patterns. 
% While these methods sometimes reproduced one specific stripe pattern (Fig. \ref{fig_result_simulated}c),
%  they failed to predict most other patterns. 
As shown in Fig.~\ref{fig_result_simulated},
 while some of the simulation results with the parameters estimated by the pooling‑based model
 successfully reproduce the specific stripe patterns (Fig.~\ref{fig_result_simulated}c), 
 most of them instead exhibit similar polka‑dot–like patterns and fail to reproduce the target patterns.
% This is because 
%  cells in these patterns are distributed across the entire area 
%  rather than being limited to specific locations. 
% Also, there is no spatial bias, such as a preference for the top, bottom, left, or right. 
This is because
 simple pooling of cell coordinates cannot capture the differences in geometric structure between cell patterns.
In these patterns,
 the cells are spread across the entire area, with little spatial concentration or bias.
% These are unique features of multicellular pattern formation.
Thus, 
 the features obtained by pooling cell coordinates show little difference between these cell patterns.
% これは、本質的に座標のプーリングでは細胞パターンの空間的構造の違いを表現することができないことに起因します。
% おおくのパターンにおいて、どの種類の細胞も空間全体にまんべんなく散らばっており、空間的偏りは少ないです。
% よって、どんな細胞パターンに対しても、プーリングによって得られる値は大差なく、座標のmin/maxプーリング値はシミュレーション空間の端付近を表し、avgプーリング値は空間の中心付近を表します。
As shown in Table~\ref{tab_combined_evaluation},
 the correlation of these methods is nearly zero, and the standard deviations are also minor. 
This indicates that
 they output the average parameters from the training data to minimize the MSE. 
In the end, these methods cannot recognize the unique features of each target. 

While PointNet and PointNet++ improved the MSE and \replaced{standard deviation ration}{standard deviation ratio} rather than Pooling GLM and MLP,
 their Pearson correlation $r$ remained near zero. 
This suggests that their predictions are distributed the same ranges as the true parameters, however the parameter 
 are originally designed to extract features, % such as edges and corners, from point clouds
 using a ``max pooling'' operation.
 % map 2D points into a high-dimensional embedding space and
 % use a "max pooling" operation to select specific features, 
 such as edges or corners.
However, as shown in Fig.~\ref{fig_result_simulated},
 cell patterns spread across the entire area and lack precise edges.
In standard point cloud tasks,
 such uniform patterns are often treated as noise.
Furthermore, 
 PointNet++ divides the space into many small regions to extract local features. 
However, for periodic cell patterns, 
 every region looks almost the same.
Therefore,
 PointNet++ receives the same information across these regions and 
 cannot extract unique features to distinguish them.
% Because the model sees the same information across regions, 
%  it cannot identify unique features to distinguish between regions. 
The low MSE of PointNet++
 does not necessarily indicate high predictive accuracy.
The near-zero correlation 
 suggests that the model cannot track actual parameter changes. 
Instead, the model likely focuses on only a specific pattern type,
 such as the dot pattern. 
By outputting parameters 
 that reflect minor differences within that particular pattern type, 
 the model keeps the MSE low.
However, it fails to recognize other patterns, such as stripes.
These limitations support
 the use of Betti vectors to capture the proper structural connectivity of multicellular patterns.

Our TI$^2$PS MLP exhibits high performance even with limited training data. In every metric, it outperforms PointNet++ (Fig.~\ref{fig_data_reduction}). 
\replaced{
For example, our framework, which uses only 10\% of the data, achieves a lower MSE than PointNet++ using 100\% of the data. 
}
{
For example, our framework, which uses only 10\% of the data, outperformed PointNet++, which used 100\% of the data, across all evaluation metrics.  
}
This remarkable advantage is important because acquiring large datasets is often expensive or restricted by ethical considerations.
The Pearson correlation of PointNet++ remains near zero, regardless of the training data size. This indicates that increasing the amount of data does not improve its performance.
The fact that the MSE of PointNet++ improves while its correlation stays at zero suggests a specific behavior. 
We hypothesize that PointNet++ only learns the range of the parameter distribution, rather than the actual relationship between patterns and values. 
The observed increase in the standard deviation ratio is consistent with this hypothesis. 
Instead of predicting the correct value for each specific input, 
PointNet++ likely generates random values based on the learned distribution. 
These weaknesses in modern architectures, such as PointNet++, highlight the advantages of our TI$^2$PS~framework for biological patterning.

\replaced{
Although the ABM simulations using the \TIIPS~MLP estimated parameters recapitulated some aspects of the target behavior, 
the model was ultimately unable to fully reproduce the mutant patterns (Fig.~\ref{fig_result_mutant}).
There are two possible explanations for this outcome.
One possibility is that the parameter combinations required to generate the mutant patterns lie outside the range of the training data.
Another possibility is that additional biological mechanisms not captured by the current simulation are involved in generating the mutant patterns.
Despite the failure to reproduce the pattern, the estimated parameters for each mutant provide insights into the causal relationships between cellular interactions and the resulting patterns. Specifically, the significant differences in the repulsion scale between different cell types ($R^{BY}$) suggest that the parameter drives the formation of the distorted stripes in \textit{dali/+} and the polka-dot patterns in \textit{leopard}.
Additionally, certain parameters, including $R^{BB}$ and $R^{YY}$ in \textit{dali/+} and $r^{BY}$ in \textit{leopard}, are close to the maximum value of the range defined in the dataset. This information is highly useful for understanding the underlying cellular mechanisms and for expanding the model or exploring parameters across a wider range of biological phenomena.
This framework may serve as a tool to uncover previously unrecognized mechanisms in multicellular interactions.
}
{
Although the ABM simulations using the parameters estimated by the \TIIPS~MLP recapitulated some aspects of the target patterns, they did not fully reproduce the manually crafted mutant patterns of \textit{dali/+} and \textit{leopard} (Fig.~\ref{fig_result_mutant}).
This discrepancy highlights a limitation of the current approach.
The current framework alone cannot determine whether the reconstruction failure is due to the limited fidelity of the hand-drawn target patterns, the target patterns lying outside the training dataset's parameter range, biological mechanisms not represented in the ABM, limitations of the inverse model, or other factors.
Therefore, this mutant-pattern experiment should be interpreted as an exploratory analysis rather than a definitive validation of the biological mechanisms underlying these mutants.
Nevertheless, the estimated parameters can still provide candidate hypotheses for further investigation.
For example, the estimated repulsion scale between different cell types ($R^{BY}$) differed substantially between the mutant patterns, suggesting that this parameter may be associated with the distorted stripes in \textit{dali/+} and the polka-dot patterns in \textit{leopard}.
In addition, certain parameters, including $R^{BB}$ and $R^{YY}$ in \textit{dali/+} and $r^{BY}$ in \textit{leopard}, were close to the upper bound of the range defined in the training dataset.
These results may help guide future analyses, such as detailed observations of intercellular interactions between different cell types and training inverse surrogate models over a wider parameter range.
}

The current results of the \TIIPS~framework suggest several directions for future research. 
\added{
Although we evaluated variability arising from the training procedure using five independent runs with different random seeds, the current framework provides point estimates and does not yet provide full parameter-wise confidence intervals or posterior uncertainty estimates.
Incorporating parameter-level uncertainty quantification, such as bootstrap-based confidence intervals, ensemble-based uncertainty estimates, or Bayesian inference, will be important for improving the reliability of mechanistic interpretation.
}\added{
Although the proposed \TIIPS~MLP showed better visual reconstruction of the global features of the target patterns than the baseline methods, discrepancies remained in certain local features.
A systematic analysis of when and why such discrepancies arise will be important for further improving topological feature extraction and inverse surrogate modeling.
}
While this study focuses on zebrafish pigment patterns, applying the framework to other complex biological systems will help confirm its broader utility. 
Future efforts will include improving data-generation efficiency, exploring topological features beyond Betti vectors, and evaluating the reliability of the surrogate model. 
These steps will help generalize the \TIIPS~framework for a wider range of multicellular systems.

\section*{Acknowledgment}
We thank Professor Melissa Kemp, Andrew’s academic supervisor, for supporting his participation in this research.

\section*{Code and Data Availability}
\added{
The code and data will be made available from the authors upon reasonable request.
}

\appendices
\section{Sensitivity Analysis of Betti-Vector Filtration Discretization}
\label{ap:betti_step}
\added{
We evaluated the sensitivity of the \TIIPS~MLP to the discretization of filtration values used to construct the Betti vectors.
Table~\ref{tab:betti_step_sensitivity} shows the results obtained using different filtration step sizes, $\epsilon_i-\epsilon_{i-1}$, while keeping the number of filtration values fixed at $N_E=1000$.
The results indicate that increasing the filtration step size improved the performance for all evaluation metrics in the tested settings.
This suggests that the range of filtration values used to construct the Betti curves influences the information captured by the resulting Betti vectors.
However, increasing the filtration step size also increased the processing time required to construct the Betti vectors.
This is because a larger filtration value allows more balls to intersect, which adds more simplices to the simplicial complex $X(\epsilon)$.
Therefore, even when the number of filtration values $N_E$ is fixed, the computational cost can increase as the filtration step size becomes larger.
Based on this sensitivity analysis, we used $\epsilon_i-\epsilon_{i-1}=0.10$ as the largest step size tested that yielded the best performance while remaining computationally feasible.
}
\begin{table}[t]
  \centering
  \caption{
  \added{
  Sensitivity analysis of Betti-vector filtration discretization. 
  The discretization step is defined as $\epsilon_i-\epsilon_{i-1}$. 
  Values are reported as the mean $\pm$ standard deviation over five independent runs with different random seeds.
  }
  }
  \label{tab:betti_step_sensitivity}
  \begin{tabular}{l | c c c}
    \toprule
    $\epsilon_i-\epsilon_{i-1}$ 
    & \makecell{MSE ($\downarrow$)} 
    & \makecell{Pearson $r$ ($\uparrow$)} 
    & \makecell{$\sigma_{\mathrm{ratio}}$ (\%) $\uparrow$)} \\
    \midrule
    0.01
    & \makecell{$0.2905$ \\ $\pm 0.0006$}
    & \makecell{$0.3627$ \\ $\pm 0.0030$}
    & \makecell{$40.9377$ \\ $\pm 0.6309$} \\

    0.05
    & \makecell{$0.2248$ \\ $\pm 0.0023$}
    & \makecell{$0.5532$ \\ $\pm 0.0062$}
    & \makecell{$60.7877$ \\ $\pm 0.9723$} \\

    0.10
    & \makecell{$\mathbf{0.1911}$ \\ $\mathbf{\pm 0.0016}$}
    & \makecell{$\mathbf{0.6360}$ \\ $\mathbf{\pm 0.0038}$}
    & \makecell{$\mathbf{68.5540}$ \\ $\mathbf{\pm 1.5241}$} \\
    \bottomrule
  \end{tabular}
\end{table}

\section{Impact of Averaging Betti Vectors}
\label{ap:betti_avg}

\added{
As described in Section~\ref{subsec:data}, the proposed \TIIPS~GLM and \TIIPS~MLP use Betti vectors averaged over five stochastic ABM realizations generated with the same parameter set.
To evaluate the impact of this averaging procedure, we compared the averaged setting with a non-averaged setting.
In the non-averaged setting, each stochastic realization was treated as an individual sample for the proposed methods.
Table~\ref{tab:betti_avg} compares the evaluation metrics of the averaged and non-averaged settings for \TIIPS~GLM and \TIIPS~MLP.
The results show that the performance remained comparable between the averaged and non-averaged Betti-vector settings.
For both \TIIPS~GLM and \TIIPS~MLP, the MSE and Pearson correlation changed only marginally when the averaging procedure was removed.
Although the $\sigma_{\mathrm{ratio}}$ showed a larger difference for \TIIPS~MLP, the overall performance trend was preserved.
These results indicate that the advantage of the proposed methods is not solely attributable to averaging over stochastic realizations.
}

\begin{table}[t]
\centering
\caption{
\added{
Impact of averaging Betti vectors for \TIIPS~GLM and \TIIPS~MLP.
Values are reported as the mean $\pm$ standard deviation over five independent runs with different random seeds.
}
}
\label{tab:betti_avg}
\begin{tabular}{l | c c c}
\toprule
Method
    & \makecell{MSE ($\downarrow$)} 
    & \makecell{Pearson $r$ ($\uparrow$)} 
    & \makecell{$\sigma_{\mathrm{ratio}}$ (\%) ($\uparrow$)} \\
\midrule
\makecell[l]{\TIIPS~GLM \\ (Averaged)}
& \makecell{$0.2476$ \\ $\pm 0.0021$}
& \makecell{$0.4933$ \\ $\pm 0.0068$}
& \makecell{$53.8072$ \\ $\pm 0.4956$} \\
\makecell[l]{\TIIPS~GLM \\ (Non-averaged)}
& \makecell{$0.2408$ \\ $\pm 0.0007$}
& \makecell{$0.5135$ \\ $\pm 0.0051$}
& \makecell{$54.4068$ \\ $\pm 0.4590$} \\
\midrule
\makecell[l]{\TIIPS~MLP \\ (Averaged)}
& \makecell{$0.1911$ \\ $\pm 0.0016$}
& \makecell{$0.6360$ \\ $\pm 0.0038$}
& \makecell{$68.5540$ \\ $\pm 1.5241$} \\
\makecell[l]{\TIIPS~MLP \\ (Non-averaged)}
& \makecell{$0.1940$ \\ $\pm 0.0031$}
& \makecell{$0.6169$ \\ $\pm 0.0039$}
& \makecell{$77.6200$ \\ $\pm 0.8681$} \\
\bottomrule
\end{tabular}
\end{table}

\bibliographystyle{IEEEtran} 
\bibliography{ref}

\end{document}